\documentclass[journal]{IEEEtran}
\usepackage{amsmath,amsfonts}
\usepackage{algorithm}
\usepackage{array}
\usepackage[caption=false,font=normalsize,labelfont=sf,textfont=sf]{subfig}
\usepackage{textcomp}
\usepackage{stfloats}
\usepackage{multirow}
\usepackage{enumitem}
\usepackage{booktabs}
\usepackage{verbatim}
\usepackage{graphicx}
\usepackage{cite}
\usepackage{xcolor}

\usepackage{url}
\usepackage{hyperref}
\hypersetup{hidelinks}
\usepackage{pdfpages}
\usepackage{comment}
\usepackage{fontawesome5}
\usepackage{amsthm} 
\usepackage{pifont}
\usepackage{cleveref}
\Crefname{figure}{Fig.}{Figs.}
\Crefname{table}{Table}{Tables}
\usepackage{bbding}
\usepackage{algpseudocode}
\usepackage{mdframed}
\usepackage{amsfonts}
\usepackage[skins,breakable]{tcolorbox}
\usepackage{tablefootnote}
\usepackage{threeparttable}
\usepackage{balance}

\newtheorem{definition}{Definition}

\newcommand{\RECTSI}{\textsc{ReCTSi}}
\newcommand{\MEAN}{\textsc{MEAN}}
\newcommand{\MF}{\textsc{MF}}
\newcommand{\TRMF}{\textsc{TRMF}}
\newcommand{\SMVNMF}{\textsc{SMV-NMF}}
\newcommand{\MICE}{\textsc{MICE}}
\newcommand{\BRITS}{\textsc{BRITS}}
\newcommand{\rGAIN}{\textsc{rGAIN}}
\newcommand{\SAITS}{\textsc{SAITS}}
\newcommand{\TimesNet}{\textsc{TimesNet}}
\newcommand{\GRIN}{\textsc{GRIN}}
\newcommand{\NET}{\textsc{NET$^3$}}

\newcommand{\PoGeVon}{\textsc{PoGeVon}}

\newcommand{\AdaCTSi}{\textsc{AdaCTSi}}

\newcommand{\AdaCTSiLTSIT}{\textsc{AdaCTSi-LTSIT}}
\newcommand{\AdaCTSiOSTCN}{\textsc{AdaCTSi-OSTCN}}
\newcommand{\AdaCTSiSSA}{\textsc{AdaCTSi-SSA}}

\newcommand{\AdaCTSionly}{\textsc{AdaCTSi-Only}}
\newcommand{\AdaCTSirand}{\textsc{AdaCTSi-Rand}}
\newcommand{\AdaCTSicw}{\textsc{AdaCTSi-CW}}
\newcommand{\AdaCTSiall}{\textsc{AdaCTSi-All}}

\begin{document}

\title{Impute On-Demand: Adaptive Correlated Time Series Imputation for Changing Environments}

\author{Zhichen Lai,
        Huan~Li$^*$,~\IEEEmembership{Member,~IEEE},
        Dalin~Zhang$^*$,~\IEEEmembership{Senior Member,~IEEE},
        Dong~Gong,
        Lina~Yao,~\IEEEmembership{Senior Member,~IEEE},
        and~Christian~S.~Jensen,~\IEEEmembership{Fellow,~IEEE},

\IEEEcompsocitemizethanks{

\IEEEcompsocthanksitem 
Z. Lai is with the College of Computer and Data Science, Fuzhou University, Fuzhou, China; the Engineering Research Center of Big Data Intelligence, Ministry of Education, Fuzhou, China; the Fujian Key Laboratory of Network Computing and Intelligent Information Processing, Fuzhou University, Fuzhou, China; and the Department of Computer Science, Aalborg University, Aalborg, Denmark. E-mail: \mbox{lzc@fzu.edu.cn}.\protect
\IEEEcompsocthanksitem D. Zhang is with the Space Information Research Institute, Hangzhou Dianzi University, Hangzhou, Zhejiang, China, and the Department of Computer Science, Aalborg University, Aalborg, Denmark. E-mail: \mbox{zhangdalin@hdu.edu.cn}.\protect
\IEEEcompsocthanksitem C. S. Jensen is with the Department of Computer Science, Aalborg University, Aalborg, Denmark. E-mail: \mbox{csj@cs.aau.dk}.\protect
\IEEEcompsocthanksitem H. Li is with the College of Computer Science and Technology, Zhejiang University, Hangzhou, China, and the State Key Laboratory of Blockchain and Data Security, Zhejiang University, Hangzhou, China. E-mail: \mbox{lihuan.cs@zju.edu.cn}.\protect
\IEEEcompsocthanksitem D. Gong is with the School of Computer Science and Engineering, The University of New South Wales, Sydney, NSW, Australia. E-mail: \mbox{dong.gong@unsw.edu.au}.\protect
\IEEEcompsocthanksitem L. Yao is with the School of Computer Science and Engineering, The University of New South Wales, Sydney, NSW, Australia, and CSIRO's Data61, Eveleigh, NSW, Australia. E-mail: \mbox{lina.yao@unsw.edu.au}.\\  *Corresponding authors: H. Li and D. Zhang.
}
}

\IEEEaftertitletext{%
\vspace{-2.85em}%
\noindent\makebox[\textwidth][c]{\parbox{0.98\textwidth}{\centering\normalfont\fontsize{5.5}{6.2}\selectfont
\textcopyright~2026 IEEE. Personal use of this material is permitted. Permission from IEEE must be obtained for all other uses, including reprinting/republishing this material for advertising or promotional purposes, creating new collective works for resale or redistribution to servers or lists, or reuse of any copyrighted component of this work in other works.\\[-0.1ex]
This article has been accepted for publication in \textit{IEEE Transactions on Knowledge and Data Engineering}. This is the author's accepted version and has not been fully edited by IEEE.}}%
\vspace{0.60em}%
}

\markboth{IEEE Transactions on Knowledge and Data Engineering,~Vol.~xx, No.~xx, xx~20xx}%
{Lai \MakeLowercase{\textit{et al.}}: Impute On-Demand: Adaptive Correlated Time Series Imputation for Changing Environments}


\maketitle

\begin{abstract}
Imputation is a well-established data-cleaning task in the database community, and deep learning has advanced the state of the art. Internet of Things (IoT) applications generate vast amounts of Correlated Time Series (CTS) data that often contain missing values and therefore require imputation.
However, current methods predominantly emphasize accuracy and often neglect the adaptability required in changing IoT environments. First, they struggle with sensor failures because CTS imputation relies on correlations among sensors; when some sensors are unavailable, the imputation of values from other sensors may be disrupted. Second, they cannot selectively impute specific sensors, leading to unnecessary processing of values known to be complete. Third, their static architectures cannot adapt to changing resource availability, resulting in consistently high computational demands.

To tackle these limitations, we propose \AdaCTSi{}, an adaptive CTS imputer for changing environments. The architecture integrates a One-shot Temporal Convolutional Network with a Learned Time-Sensor Index Table to extract and decouple complex spatio-temporal features into sensor-wise embeddings, thereby enabling adaptation to varying sensor subsets. Sparse Spatial Attention extracts dynamic spatial correlations efficiently, while Correlation-Weighted Sensor Selection ensures sufficient spatial context by selecting informative sensors.
Experimental results across twelve baseline methods, three adaptability scenarios, and five benchmark datasets spanning traffic, air quality, and trajectory domains show that \AdaCTSi{} achieves an average 33.1\% reduction in MAE relative to the strongest baseline on each dataset, while supporting sensor-subset and resource-adaptive inference with a single trained model. Moreover, its modest memory footprint enables deployment on a range of commodity computing devices, including MCUs, making \AdaCTSi{} an adaptive CTS imputation solution for changing environments. The source code is available at \url{https://github.com/ryanlaics/adactsi/}.

\end{abstract}


\begin{figure}[!htbp]
    \begin{center}
        \includegraphics[width=\linewidth]{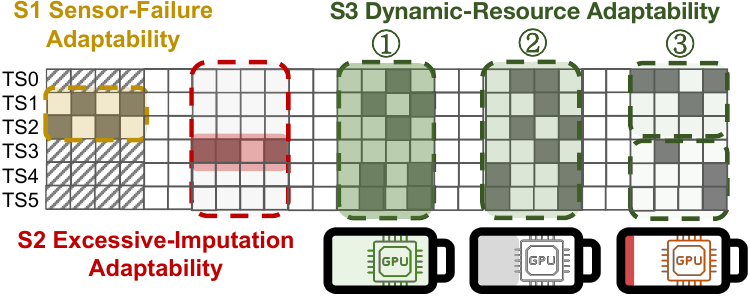}
    \end{center}
    \caption{Adaptability scenarios for CTS imputation: Sensor-Failure Adaptability (S::\texttt{{SFA}}) excludes failed sensors; Excessive-Imputation Adaptability (S::\texttt{{EIA}}) selectively processes incomplete sensors; and Dynamic-Resource Adaptability (S::\texttt{{DRA}}) adjusts the model and sensor-group size to available resources.}
    \label{fig:adp}
\end{figure}
\section{Introduction}
\label{sec:Introduction}
Internet of Things (IoT) applications, including traffic management~\cite{zhao2023multiple,lai2024lightcts}, wearable devices~\cite{lai2024e2usd,wang2023time2state}, and database monitoring systems~\cite{yao2023simplets,mohr2024hardware}, rely on interconnected sensors to monitor operational parameters such as traffic speed in road networks.
These sensors generate vast amounts of highly correlated time series data because of their spatio-temporal relationships; we refer to such data as Correlated Time Series (CTS). CTS data are often stored in spatio-temporal or temporal databases~\cite{khelifati2023tsm,wang2020apache} for downstream tasks, including classification~\cite{campos2023lightts,lai2024e2usd}, anomaly detection~\cite{he2023oneshotstl,campos2021unsupervised}, and forecasting~\cite{lai2024lightcts,chen2023multi}.
In practice, sensor failures, network interruptions, and battery depletion introduce missing values that distort CTS patterns and degrade downstream analyses, as shown by the forecasting results in \Cref{tab:forecasting_results}. The resulting false alerts or biased predictions~\cite{blazquez2023selective} motivate accurate CTS imputation.

Recent deep learning (DL) imputers achieve state-of-the-art performance~\cite{tashiro2021csdi,liu2023pristi,wang2023networked,yuan2024nuhuo,rectsi2024resource,du2023saits,li2023missing}, but typically assume fixed structures, capacities, and sensor partitions. In deployment, available FLOPs and memory vary, as does the workload\footnote{The workload is the number of input sensors because DL-based imputers typically use short, fixed-size windows (e.g., 24 time steps).}. Complete values need no imputation, while failed or removed sensors should be excluded because their placeholders can corrupt correlation-based estimates for other sensors.

Recent practices in \emph{elastic computing}~\cite{li2022serverless, fouladi2017encoding, li2022serverlessacm, shen2022resource} motivate adaptive CTS imputation frameworks that adjust model configurations to online workloads and available resources. Predictive resource provisioning and scheduling provide complementary system-level mechanisms~\cite{poppe2020seagull, bashir2021take, shen2022resource}; this paper focuses on the imputation model design and addresses the following question:

\smallskip
\textit{Is it possible to design a DL-based imputer (i.e., imputation model) that can adapt to variations in online workload and available computing resources to perform \textbf{on-demand inference}, thereby improving resource utilization while maintaining imputation accuracy?}
\smallskip

The on-demand inference capability of such an imputer can support at least three desirable scenarios:
\begin{itemize}[labelindent=0pt,labelwidth=1.4em,labelsep=0.5em,leftmargin=1.9em,topsep=0.35em,itemsep=0.35em]
\item[S1] \textbf{Sensor-Failure Adaptability} (S::\texttt{{SFA}}). 
As depicted in the yellow sector of \Cref{fig:adp}, failed sensors (i.e., TS0 and TS3--TS5) can significantly disrupt the imputation of values from other sensors because imputation relies on learning correlations among sensors. Existing DL-based imputers assume a constant input dimensionality, so when sensors fail, they often fill the unavailable sensor inputs with default or random values. These placeholder values can reduce the accuracy of imputing the remaining sensors and incur unnecessary computation. Hence, an imputer should adjust its input size and exclude failed sensors. By reducing the input dimensionality on the fly, the model avoids erroneous placeholder values and reduces computational demand. This dynamic, on-demand inference capability enables stable operation during sensor failures, maintenance activities, and changing online workloads, enhancing robustness and efficiency. We study this scenario experimentally in \Cref{subsec:sfada}.

\item[S2] \textbf{Excessive-Imputation Adaptability}  (S::\texttt{{EIA}}).
The red sector of \Cref{fig:adp} illustrates that when most sensors report complete data, current DL-based methods typically process all sensors uniformly and cannot easily restrict computation to the incomplete sensors (e.g., TS3), leading to unnecessary processing and excessive computation.
This inefficiency is exacerbated in large-scale sensor settings, where only a few sensors have missing values. Inspired by dynamic resource allocation in elastic computing~\cite{li2022serverless}, there is a need for imputation approaches that can reduce the input workload while still leveraging informative complete sensors. Such selective processing minimizes unnecessary computation and allocates resources to the most informative signals, improving both efficiency and imputation quality. We study this scenario in \Cref{subsec:eiada}.

\item[S3] \textbf{Dynamic-Resource Adaptability}  (S::\texttt{{DRA}}).
The green sectors in \Cref{fig:adp} illustrate that the static structures and parameter capacities of current DL-based imputers impose consistently high computational demands and limit adaptation to changing online workloads and available resources. An adaptive imputer can instead match its computational requirements to the available resources in real time. When resources are constrained, it can reduce demand by simplifying its architecture (\ding{172} to \ding{173}, deactivating some parameters) and partitioning the input (\ding{173} to \ding{174}, splitting sensors into the groups TS0--TS2 and TS3--TS5). Conversely, when more resources become available, it can activate additional parameters and increase the partition size to process more or all sensors simultaneously. This adaptability can improve resource utilization and imputation performance under online conditions. We study this scenario experimentally in \Cref{subsec:drada}.
\end{itemize}

Addressing these scenarios requires methods that go beyond static structures and parameter capacities by supporting variable sensor subsets and input dimensionalities.

\begin{figure}[ht]
\centering
        \includegraphics[width=0.55\linewidth]{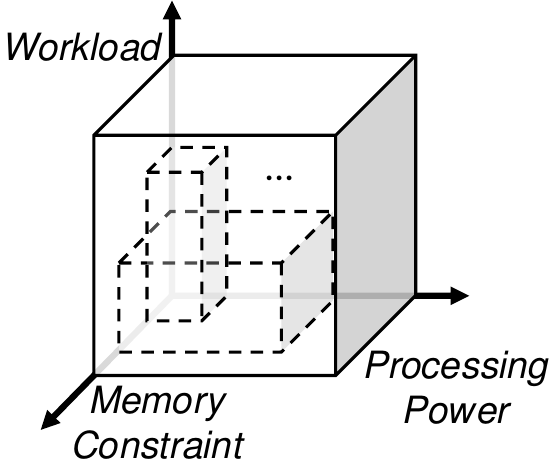}
	\caption{On-demand Mode for Handling Variations in Input Workload and Resource Availability.}
	\label{fig:mode}
\end{figure}


As shown in \Cref{fig:mode}, \emph{On-demand Mode} uses one adaptable $N$-in-1 model and a shared parameter set across configurations, avoiding the design, training, deployment, and maintenance of many workload- or resource-specific models.

We therefore present \AdaCTSi{}, an adaptive CTS imputer for on-demand inference. Its \emph{One-shot Temporal Convolutional Network} (One-shot TCN) and \emph{Learned Time-Sensor Index Table} (LTSIT) produce sensor-wise spatio-temporal representations. \emph{Sparse Spatial Attention} (SSA) extracts informative correlations from variable sensor subsets, while \emph{Correlation-Weighted Sensor Selection} (CW-SS) supplements incomplete sensors with informative complete ones when needed.
The key contributions are summarized as follows.
\begin{itemize}[leftmargin=*]
    \item We identify three practical adaptability challenges in CTS imputation under changing workloads and computing resources.
    \item We propose \AdaCTSi{}, which combines One-shot TCN, LTSIT, SSA, and CW-SS to support variable sensor subsets, model capacities, and resource budgets.
    \item Extensive experiments, including downstream forecasting, show that \AdaCTSi{} supports on-demand adaptability while consistently achieving higher accuracy than existing imputers.
\end{itemize}

The remainder of this paper presents the preliminaries, design principles, method, experiments, related work, and conclusion.

\section{Preliminaries}
\label{sec:preliminaries}
\begin{table*}[htbp]
\centering
\caption{Design Principles (P1 to P3) for Adaptability in Different Scenarios (S1 to S3 in \Cref{sec:Introduction})}
\resizebox{\textwidth}{!}{
\begin{tabular}{lccc}
\toprule
\textbf{Design Principles} & \textbf{P1. Sensor Variation Handling} & \textbf{P2. Informative Sensor Selection} & \textbf{P3. Model Complexity Tuning} \\ 
\midrule
{S1. Sensor-Failure Ada.} & \faStar[regular]~\textit{Adapt to available sensors} & Not required &  Not required \\ 
{S2. Excessive-Imputation Ada.} & \faStar[regular]~\textit{Process only sensors with missing data} & \faStar[regular]~\textit{Select sensors for spatial information} & Not required \\ 
{S3. Dynamic-Resource Ada.} & \faStar[regular]~\textit{Partition sensors to match memory capacity} &  Not required & \faStar[regular]~\textit{Adapt complexity to processing power} \\ 
\midrule
{Existing DL-based Imputers} & \faThumbsDown[regular]~Unsupported & \faThumbsDown[regular]~Unsupported & \faThumbsDown[regular]~Unsupported \\ 
\textbf{\AdaCTSi{} (ours)} & \faThumbsUp[regular]~Supported (One-shot TCN, LTSIT, SSA, CW-SS) & \faThumbsUp[regular]~Supported (SSA, CW-SS) & \faThumbsUp[regular]~Supported (SSA) \\ 
\bottomrule
\end{tabular}
}
\label{tab:design_principles}
\end{table*}

We formalize the problem of Correlated Time Series (CTS) imputation and introduce key definitions that form the foundation of our adaptive framework.

\begin{definition}[Correlated Time Series, CTS]
\emph{Correlated Time Series} (CTS) are multiple time series generated by sensors monitoring a shared set of processes. Consider a set of $\mathtt{N}$ sensors, each producing a time series. Let $\mathbf{X} \in \mathbb{R}^{\mathtt{N} \times \mathtt{T}}$ denote these time series, where $\mathbf{X}_{i,t}$ is the value from the $i$-th sensor at time $t$~\cite{wu2021autocts}. 
In CTS, \textbf{temporal correlations} refer to the dependencies along the time axis within a single sensor's series; for example, the measurement at time $t$ can be influenced by measurements at previous time steps. In contrast, \textbf{spatial correlations} denote dependencies among measurements from different sensors, observed at the same time or even across time, because these sensors monitor shared or related processes\footnote{Cross-sensor correlations can reflect more than spatial proximity (e.g., functional or process-based dependencies), but following convention, we refer to them as spatial correlations~\cite{lai2023lightcts}.}. Following prior work (e.g.,~\cite{wu2021autocts}), we do not impose a fixed correlation threshold; correlations can be positive or negative and are dynamic and application-dependent in practice.
\end{definition}

\begin{definition}[CTS Imputation]
Given a CTS $\mathbf{X} \in \mathbb{R}^{\mathtt{N} \times \mathtt{T}}$ with missing values indicated by a mask matrix $\mathbf{M} \in \{0,1\}^{\mathtt{N} \times \mathtt{T}}$ (where "$0$" denotes missing), \emph{CTS imputation} estimates the missing values to improve data quality. The imputation process replaces only missing values.
\begin{equation}
\hat{\mathbf{X}} = \mathbf{X} \odot \mathbf{M} + \mathit{g} (\mathbf{X} \mid \mathbf{M}) \odot (\mathbf{1} - \mathbf{M}),
\end{equation}
where $\hat{\mathbf{X}} \in \mathbb{R}^{\mathtt{N} \times \mathtt{T}}$ is the imputed CTS, $\odot$ denotes element-wise multiplication, $\mathbf{1} \in \mathbb{R}^{\mathtt{N} \times \mathtt{T}}$ represents a matrix of ones, and $\mathit{g}(\mathbf{X} \mid \mathbf{M})$ is the imputation function that estimates the missing values based on the observed data $\mathbf{X}$ and the mask $\mathbf{M}$.
\end{definition}

\begin{definition}[Adaptive CTS Imputation]
\emph{Adaptive CTS imputation} refers to a model's ability to adjust its architecture and input processing in response to changing workloads and resource availability to ensure accurate, on-demand imputation. The required forms of adaptability include Sensor-Failure Adaptability, Excessive-Imputation Adaptability, and Dynamic-Resource Adaptability.
Specifically, an imputer should be able to exclude failed sensors, selectively process incomplete subsets of sensors, and adapt its resource requirements by adjusting both its architecture and the size of input sensor partitions on demand. The adaptive imputation process is defined as follows.
\begin{equation}
\hat{\mathbf{X}}_{\mathcal{S}}^{c} = \mathbf{X}_{\mathcal{S}} \odot \mathbf{M}_{\mathcal{S}} + \mathit{g}^c\left(\mathbf{X}_{\mathcal{S}} \mid \mathbf{M}_{\mathcal{S}}\right) \odot \left(\mathbf{1} - \mathbf{M}_{\mathcal{S}}\right),
\end{equation}
where $\mathcal{S} \subseteq \{1, 2, \dots, \mathtt{N}\}$ represents the subset of sensors being processed and $\mathbf{X}_{\mathcal{S}} \in \mathbb{R}^{|\mathcal{S}| \times \mathtt{T}}$ contains the data from this subset. $\mathbf{M}_{\mathcal{S}} \in \{0, 1\}^{|\mathcal{S}| \times \mathtt{T}}$ is the corresponding mask matrix. The imputation function $\mathit{g}^c(\mathbf{X}_{\mathcal{S}} \mid \mathbf{M}_{\mathcal{S}})$ estimates the missing values, with model complexity level $c$ determining its resource demands. The matrix $\mathbf{1}$ has the same shape as $\mathbf{M}_{\mathcal{S}}$. Finally, $\hat{\mathbf{X}}_{\mathcal{S}}^{c}$ denotes the imputed CTS for the selected sensors under input size $|\mathcal{S}|$ and complexity level $c$.
\end{definition}

\begin{figure}[ht]
	\begin{center}
		\includegraphics[width=\linewidth]{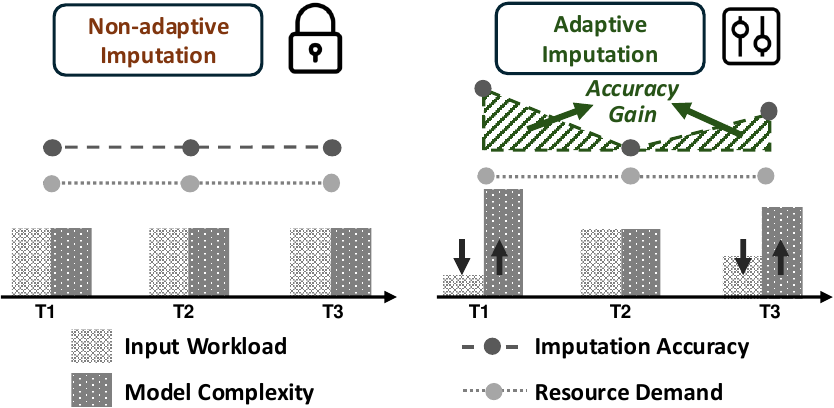}
	\end{center}
	\caption{Non-adaptive vs.\ Adaptive Imputation under Stable Resource Demand.}
	\label{fig:2d}
\end{figure}

\Cref{fig:2d} illustrates how the imputer adjusts its complexity to the input workload to maintain stable resource demand. In this scenario, the left diagram represents non-adaptive imputation, where both the input workload and model complexity remain constant at time points $T1$, $T2$, and $T3$, resulting in a relatively stable accuracy level.
In contrast, the right diagram shows adaptive imputation, in which the imputer dynamically increases its complexity in response to lower workloads at $T1$ and $T3$, thereby improving accuracy under a fixed resource budget. The shaded areas in \Cref{fig:2d} highlight this accuracy gain.

\section{Design Principles}
\label{sec:design}
As discussed in \Cref{sec:Introduction}, CTS imputation in real-world applications encounters three primary challenges: Sensor-Failure Adaptability (S::\texttt{SFA}), Excessive-Imputation Adaptability (S::\texttt{EIA}), and Dynamic-Resource Adaptability (S::\texttt{DRA}). While existing DL-based approaches have achieved strong CTS imputation performance, they assume a fixed model structure and parameter capacity during deployment. This rigidity causes inefficiency and performance degradation when sensor availability, imputation demand, or computing resources change.

To broaden the applicability of imputation, we propose three key design principles, each addressing a specific challenge and guiding adaptation to changing conditions. Together, these principles form the foundation of \AdaCTSi{} for adaptive CTS imputation.

\begin{itemize}[leftmargin=*]
\item[P1] \textbf{Sensor Variation Handling} (P::\texttt{SVH}).
In IoT applications, the available sensor set may vary because of failures, maintenance activities, or changing online workloads. Moreover, in many practical scenarios, only a small subset of sensors contains missing data, while the remaining sensors provide complete data. Existing DL-based imputers nevertheless process all sensors uniformly, including failed or unnecessary ones, which can degrade imputation performance and increase computational overhead. P::\texttt{SVH} addresses this limitation by selectively processing relevant sensors. To achieve such adaptability, an imputer should decouple complex spatio-temporal features into sensor-wise representations, enabling effective imputation under sensor variation.

\item[P2] \textbf{Informative Sensor Selection} (P::\texttt{ISS}). 
Traditional DL-based imputers process data from all sensors regardless of their characteristics, leading to excessive and inefficient computation. P::\texttt{ISS} mitigates this problem by identifying the most informative sensors and using their data for spatial feature extraction. To achieve P::\texttt{ISS}, an imputer should adopt metrics such as sensor-wise similarity to evaluate each sensor's informativeness. Focusing on key sensors can reduce computation while preserving imputation accuracy.
    
\item[P3] \textbf{Model Complexity Tuning} (P::\texttt{MCT}).
Deploying imputation models across diverse environments requires models with adjustable computational demands. Existing DL-based imputers maintain a constant model structure and parameter capacity, resulting in consistently high computational demands. To achieve P::\texttt{MCT}, an imputer should support on-demand complexity tuning based on available resources. By dynamically activating subsets of model parameters or selectively processing feature maps, an imputer can scale its computational load and operate across varying resource levels.
\end{itemize}

As illustrated in \Cref{tab:design_principles}, an imputer that adopts the three design principles can adapt to sensor variation, prioritize informative sensors for effective imputation, and adjust computational demand according to available resources.

\section{\AdaCTSi{} Design}
\label{sec:AdaCTS}
\begin{figure*}[ht]
	\begin{center}
		\includegraphics[width=\linewidth]{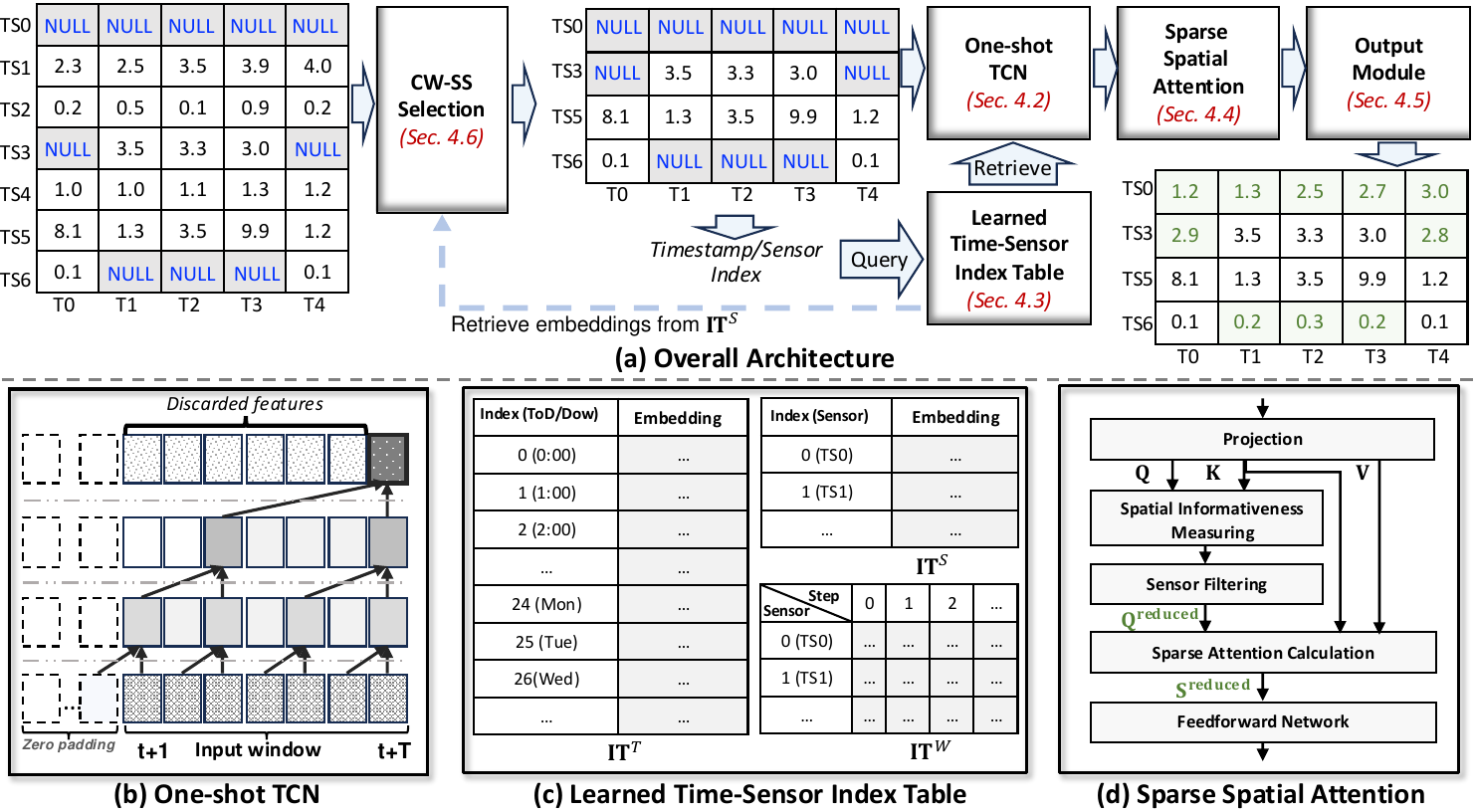}
	\end{center}
		\caption{(a) Overview of \AdaCTSi{}. CW-SS optionally selects input sensors when the set of incomplete sensors is too small to provide adequate correlation information. (b) One-shot TCN then extracts and compresses temporal features, which are enriched with multi-view spatio-temporal embeddings from (c) LTSIT. Next, (d) the SSA module applies self-attention to the most informative sensors, capturing dynamic spatial correlations while improving computational efficiency. Finally, the output module independently generates imputed values for each sensor.}
	\label{fig:all}
\end{figure*}

Building on the previously discussed design principles, we introduce \AdaCTSi{}, an adaptive data imputer for CTS. \AdaCTSi{} is specifically designed to address Sensor-Failure Adaptability (S::\texttt{{SFA}}), Excessive-Imputation Adaptability (S::\texttt{{EIA}}), and Dynamic-Resource Adaptability (S::\texttt{{DRA}}).

\subsection{The Overall Architecture of \textnormal{\AdaCTSi{}}}
\label{ssec:overall}

As depicted in \Cref{fig:all}(a), \AdaCTSi{} comprises several key components that jointly enable on-demand imputation across dynamic scenarios:
\begin{itemize}[leftmargin=*]
    \item {The input data are first processed by the \textbf{Correlation-Weighted Sensor Selection (CW-SS)} mechanism, which selects the most spatially informative sensors for on-demand imputation when the set of incomplete sensors is too small to provide sufficient spatial correlation information (P::\texttt{ISS}).}
    
    \item {Next, the \textbf{One-shot Temporal Convolutional Network (One-shot TCN)} processes the selected sensor data, as shown in \Cref{fig:all}(b). This module extracts temporal features and employs a one-shot mechanism to reduce temporal dimensionality. It decouples complex spatio-temporal features into one-dimensional sensor-wise representations (P::\texttt{SVH}), allowing subsequent spatial modules to operate without managing temporal variations.}

    \item {Subsequently, the \textbf{Learned Time-Sensor Index Table (LTSIT)} module (\Cref{fig:all}(c)) retrieves learned embeddings through multi-view indexing. This module captures periodic temporal patterns, sensor-specific characteristics, and local intra-window features via table lookups, thereby constructing comprehensive sensor-wise latent representations (P::\texttt{SVH}).}
    
    \item {Finally, as demonstrated in \Cref{fig:all}(d), the \textbf{Sparse Spatial Attention (SSA)} module captures spatial correlations without channel mixing, preserving distinct sensor features and preventing sensor-wise feature fusion. The SSA adjusts its attention sparsity rate to enable P::\texttt{MCT}, focusing on data from the most spatially informative sensors for more effective spatial correlation extraction (P::\texttt{ISS}).
    The output module then generates the final imputed data.}




\end{itemize}
Together, these four components enable adaptive CTS imputation. We describe CW-SS last because it depends on the embeddings generated by LTSIT.

\subsection{One-shot Temporal Convolutional Network}
As depicted in \Cref{fig:all}(b), the One-shot TCN extracts and compresses temporal features for each sensor by transforming input feature maps of size $\mathtt{N} \times \mathtt{T} \times \mathtt{D}$ (number of sensors $\times$ window length $\times$ embedding size) into a compact $\mathtt{N} \times \mathtt{D}$ representation. This compression decouples complex spatio-temporal features into sensor-wise one-dimensional vector representations and simplifies the design of subsequent spatial modules, which no longer need to account for temporal variations, thereby supporting the P::\texttt{SVH} principle.

The One-shot TCN employs dilated causal convolutions~\cite{DBLP:journals/corr/YuK15} to capture both short- and long-term temporal patterns within the input window. The input, denoted as $\mathbf{E}^{I} \in \mathbb{R}^{\mathtt{N} \times \mathtt{T} \times (2+\mathtt{d}_w)}$, comprises the concatenation of the input CTS $\mathbf{X} \in \mathbb{R}^{\mathtt{N} \times \mathtt{T}}$, the mask matrix $\mathbf{M} \in \mathbb{R}^{\mathtt{N} \times \mathtt{T}}$, and the intra-window embedding $\mathbf{E}^W \in \mathbb{R}^{\mathtt{N} \times \mathtt{T} \times \mathtt{d}_w}$ from the LTSIT (discussed in \Cref{subsubsec:LTSIT}). This input is first embedded into a latent representation $\mathbf{H} \in \mathbb{R}^{\mathtt{N} \times \mathtt{T} \times \mathtt{d}_d}$ using a CNN embedding layer:
    \begin{equation}
    \mathbf{H}[i, t, d] = \text{ReLU} \left(\mathbf{E}^{I}[i, t, :]\mathbf{W}_d + \mathbf{b}_d \right),
    \end{equation}
    where $\mathbf{W}_d \in \mathbb{R}^{\mathtt{k}_e \times (2+\mathtt{d}_w) \times \mathtt{d}_d}$ is the convolutional filter, $\mathbf{b}_d \in \mathbb{R}^{\mathtt{d}_d}$ is the bias term, and $\mathtt{k}_e$ is the kernel size.

Next, the TCN is applied:
\begin{equation*}
\begin{aligned}
\operatorname{TCN}(\mathbf{H} \mid \delta, \mathtt{k}_t) &= \mathbf{H}', \quad \text{where} \\
\mathbf{H}'[i, t, d] &= \sum_{p=0}^{\mathtt{k}_t-1} \left( \mathbf{H}[i, t - \delta \times p, :] \cdot W^d[p, :] \right),
\end{aligned}
\end{equation*}
where $\mathbf{H}' \in \mathbb{R}^{\mathtt{N} \times \mathtt{T} \times \mathtt{d}_d}$ is the latent representation of the TCN, $W^d[p, :] \in \mathbb{R}^{\mathtt{d}_d}$ is the $d$-th convolutional filter at position $p$, $\mathtt{k}_t$ is the kernel size, and $\delta$ is the dilation rate.

A one-shot mechanism is then applied, retaining only the feature at the final time step:
\begin{equation}
    \mathbf{E}^D[i, d] = \mathbf{H}'[i, -1, d].
    \end{equation}

The resulting compressed representation $\mathbf{E}^D \in \mathbb{R}^{\mathtt{N} \times \mathtt{d}_d}$ encapsulates each sensor's temporal features in a more manageable and compact form.

\subsection{Learned Time-Sensor Index Table}
\label{subsubsec:LTSIT}
Inspired by decoupled learning~\cite{rectsi2024resource,qian2022static}, we propose a \textbf{Learned Time-Sensor Index Table (LTSIT)} to extract and decouple spatio-temporal patterns from multiple views, as shown in \Cref{fig:all}(c). The LTSIT supports P::\texttt{SVH} by enabling on-demand embedding retrieval with constant time complexity ($\mathcal{O}(1)$) via table lookups. These embeddings are learned from the training data and indexed by sensor, timestamp, or input-window position. Once trained, the LTSIT directly retrieves multi-view feature representations during inference without additional feature-extraction computation. This design makes it highly adaptable to sensor variation (P::\texttt{SVH}).

{Unlike prior work that typically handles either temporal or spatial patterns, LTSIT's key innovation lies in its joint learning of \textit{periodic temporal}, \textit{sensor-specific}, and \textit{intra-window} patterns through three specialized index tables.} 

\begin{itemize}[leftmargin=*]
    \item \textbf{{Periodic Temporal} Index Table} ($\mathbf{IT}^T$): Captures \textit{periodic temporal patterns}. Each unique temporal key $\mathbf{k}^T$, formed by periodic indices such as Day of Week (DoW) and Time of Day (ToD) from the input window's first timestamp, is mapped to a corresponding embedding $\mathbf{v}^T \in \mathbb{R}^{\mathtt{d}_t}$. This embedding is broadcast across all sensors. Formally, the temporal embeddings are represented as:

    \begin{equation}
        \mathbf{IT}^T: \{\mathbf{k}^T \rightarrow \mathbf{v}^T \mid \mathbf{k}^T = [\text{DoW}, \text{ToD}, \ldots],  \mathbf{v}^T \in \mathbb{R}^{\mathtt{d}_t}\},
    \end{equation}
    \begin{equation}
        \mathbf{E}^T = [\mathbf{v}^T]^{\times \mathtt{N}}, \quad \mathbf{E}^T \in \mathbb{R}^{\mathtt{N} \times \mathtt{d}_t}.
    \end{equation}
    Here, $\mathbf{E}^T$ represents the temporal embedding, with each sensor sharing the same embedding $\mathbf{v}^T$ derived from $\mathbf{IT}^T$.

    \item \textbf{{Sensor Identity} Index Table} ($\mathbf{IT}^S$): Encodes \textit{sensor{-specific} patterns} by assigning each sensor identity to a unique embedding $\mathbf{v}_n^S \in \mathbb{R}^{\mathtt{d}_s}$, capturing inherent characteristics of different sensors:
    \begin{equation}
        \mathbf{IT}^S: \{\mathbf{k}^S \rightarrow \mathbf{v}_n^S \mid \mathbf{k}^S \in \{1, \ldots, \mathtt{N}\}, \quad \mathbf{v}_n^S \in \mathbb{R}^{\mathtt{d}_s}\},
    \end{equation}
    resulting in the spatial embedding:
    \begin{equation}
        \mathbf{E}^S = [[\mathbf{v}_{n}^S]^\top \mid n=1 \to \mathtt{N}], \quad \mathbf{E}^S \in \mathbb{R}^{\mathtt{N} \times \mathtt{d}_s}.
    \end{equation}

    \item \textbf{{Spatio-Temporal Window} Index Table} ($\mathbf{IT}^W$): Captures \textit{local intra-window interactions} by assigning a position-aware embedding $\mathbf{v}_{n,t}^W \in \mathbb{R}^{\mathtt{d}_w}$ to each data point $\mathbf{X}_{n,t}$ in the input window. This allows \AdaCTSi{} to learn patterns across sensors and time steps within the window. Formally, the window embeddings are represented as:
    \begin{equation}
    \begin{aligned}
        \mathbf{IT}^W\!:\! \{\mathbf{k}^W \!\rightarrow\! \mathbf{v}_{n,t}^W \mid
        \mathbf{k}^W \!=\! (n\!=\!1 \!\to\! \mathtt{N}, t\!=\!1 \!\to\! \mathtt{T}), \\
        \mathbf{v}_{n,t}^W \!\in\! \mathbb{R}^{\mathtt{d}_w}\},
    \end{aligned}
    \end{equation}
    leading to the window embedding:
    \begin{equation}
        \mathbf{E}^W \!=\! [[\mathbf{v}_{n,t}^W \mid t\!=\!1 \!\to\! \mathtt{T}] \!\mid\! n \!=\! 1 \!\to\!\mathtt{N}], \mathbf{E}^W \!\in\! \mathbb{R}^{\mathtt{N} \times \mathtt{T} \times \mathtt{d}_w}.
    \end{equation}
\end{itemize}

The window embedding $\mathbf{E}^W$ is concatenated with the input CTS $\mathbf{X}$ and the mask matrix $\mathbf{M}$ to form the input for the One-shot TCN. Subsequently, the embeddings $\mathbf{E}^T$ and $\mathbf{E}^S$ are concatenated with the output of the One-shot TCN, $\mathbf{E}^D$, to create a comprehensive feature representation:
\begin{equation}
\mathbf{E} = \text{Concat}(\mathbf{E}^D, \mathbf{E}^S, \mathbf{E}^T), \quad \mathbf{E} \in \mathbb{R}^{\mathtt{N} \times \mathtt{D}},
\end{equation}
where $\mathtt{D} = \mathtt{d}_d + \mathtt{d}_s + \mathtt{d}_t$.


{The concatenated representation $\mathbf{E}$ combines compressed temporal features with periodic temporal, sensor-specific, and intra-window embeddings, delivering a streamlined sensor-wise feature representation to subsequent modules. Specifically, LTSIT's constant-time retrieval mechanism facilitates sensor-agnostic feature extraction. This approach also supports the P::\texttt{SVH} principle by effectively addressing sensor variation.}

\subsection{Sparse Spatial Attention}
While some existing studies adopt self-attention for spatial correlation extraction~\cite{wang2023networked,du2023saits,rectsi2024resource}, they assume a constant input sensor dimensionality, limiting their adaptability to scenarios with sensor variation (P::\texttt{SVH}). Even if failed-sensor inputs are filled with zeros, these methods suffer from degraded imputation accuracy and incur unnecessary computational costs, as evidenced by \Cref{tab:drift}.

\begin{figure*}[ht]
    \begin{center}
        \includegraphics[width=\linewidth]{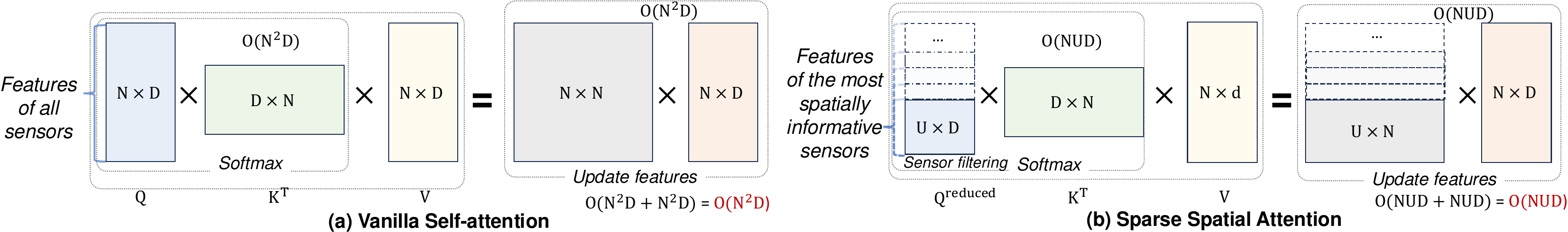}
    \end{center}
    \caption{Computational Processes of (a) Vanilla Self-Attention~\cite{vaswani2017attention} and (b) Sparse Spatial Attention. {Here, $\mathtt{U}$ is the number of sensors selected from the $\mathtt{N}$ total sensors and represents a sparsity level tied to processing power, $\mathtt{D}$ is the embedding dimension, and $\mathcal{O}(\cdot)$ denotes time complexity.}
}
    \label{fig:self-att}
\end{figure*}

To address this limitation, we introduce the \textbf{Sparse Spatial Attention (SSA)} module, designed to capture dynamic spatial correlations among sensors under varying input conditions. As shown in \Cref{fig:self-att}, unlike traditional self-attention mechanisms that require the complete sensor set, SSA performs adaptive sparse self-attention solely on the sensor subset, without channel mixing. This design avoids sensor-wise feature fusion, enabling the module to adapt to sensor variation in line with the P::\texttt{SVH} principle. By employing a sparse attention mechanism, SSA identifies and prioritizes the most spatially informative sensors, enhancing both imputation efficiency and accuracy (P::\texttt{ISS}). Moreover, SSA can adjust its sparsity level based on available processing power, enabling the imputer to scale its computational demands and operate across devices with different processing capacities (P::\texttt{MCT}).

The SSA module follows these steps:
\begin{enumerate}[leftmargin=*]
    \item \textbf{Self-Attention Projection}: Computes the queries $\mathbf{Q}$, keys $\mathbf{K}$, and values $\mathbf{V}$ from the combined feature $\mathbf{E} \in \mathbb{R}^{\mathtt{N} \times \mathtt{D}}$:
    \begin{equation}
    \mathbf{Q} = \mathbf{E} \mathbf{W}_{Q}, \quad \mathbf{K} = \mathbf{E} \mathbf{W}_{K}, \quad \mathbf{V} = \mathbf{E} \mathbf{W}_{V},
    \end{equation}
    where $\mathbf{W}_{Q}, \mathbf{W}_{K}, \mathbf{W}_{V} \in \mathbb{R}^{\mathtt{D} \times \mathtt{D}}$ are learned projection matrices.

    \item \textbf{Spatial Informativeness Measurement}: Measures the spatial informativeness of each sensor to identify the most informative ones (P::\texttt{ISS}). For a query $q_i$ and keys $\mathbf{K}$, the measure $\mathbf{M}(q_i, \mathbf{K})$ is defined as:
    \begin{equation}
    \mathbf{M}(q_i, \mathbf{K}) = {\ln  \sum_{j=1}^{\mathtt{N}} e^{\frac{q_i k_j^T}{\sqrt{\mathtt{D}}}} - \frac{1}{\mathtt{N}} \sum_{j=1}^{\mathtt{N}} \frac{q_i k_j^T}{\sqrt{\mathtt{D}}},}
    \end{equation}
     {where the first term measures the compatibility between query $q_i$ and all keys $k_j$. The exponential emphasizes strongly aligned query-key pairs, while the logarithm controls the scale and prevents large values from dominating. The second term subtracts the mean interaction, smoothing the measure and enabling more balanced comparisons among sensors.}
     
    \item \textbf{Sensor Filtering}: During training, a random sparsity scheme selects a subset of $\mathtt{U}$ sensors from the $\mathtt{N}$ sensors, where $\mathtt{U} \leq \mathtt{N}$ is a randomly chosen integer corresponding to the sparsity rate. During inference, $\mathtt{U}$ is set according to the computational requirements, and the top $\mathtt{U}$ sensors according to $\mathbf{M}(q_i, \mathbf{K})$ are selected for self-attention.

    \item \textbf{Sparse Attention Calculation}: Performs self-attention on the filtered sensors to compute attention scores and derive output features $\mathbf{A}$:
    \begin{equation}
    \mathbf{A}(\mathbf{Q}^{\text{reduced}}, \mathbf{K}, \mathbf{V}) = \text{Softmax}\left( \frac{\mathbf{Q}^{\text{reduced}} \mathbf{K}^\top}{\sqrt{\mathtt{D}}} \right) \mathbf{V},
    \end{equation}
    where $\mathbf{Q}^{\text{reduced}} \in \mathbb{R}^{\mathtt{U} \times \mathtt{D}}$ denotes the queries of filtered sensors and $\mathbf{K}, \mathbf{V} \in \mathbb{R}^{\mathtt{N} \times \mathtt{D}}$ are the keys and values, respectively.

    \item \textbf{Add \& Norm and Feed-Forward Network}: Applies residual connections and layer normalization:
    \begin{equation}
    \mathbf{Z} = \text{LayerNorm}(\mathbf{A} + \mathbf{E}).
    \end{equation}
    The module then passes the result through the feed-forward network for nonlinear transformation:
    \begin{equation}
    \text{FFN}(\mathbf{Z}) = \text{LayerNorm}(\text{ReLU}(\mathbf{Z} \mathbf{W}_1 + \mathbf{b}_1) \mathbf{W}_2 + \mathbf{b}_2 + \mathbf{Z}),
    \end{equation}
    where $\mathbf{W}_1, \mathbf{W}_2 \in \mathbb{R}^{\mathtt{D} \times \mathtt{D}}$ are weight matrices and $\mathbf{b}_1, \mathbf{b}_2 \in \mathbb{R}^{\mathtt{D}}$ are bias terms.

\end{enumerate}

The SSA module enhances adaptability in CTS imputation by accommodating varying numbers of input sensors (P::\texttt{SVH}). By using sparse spatial attention to focus on the most informative sensors (P::\texttt{ISS}), SSA robustly extracts spatial correlations even from a reduced sensor set. In addition, the random sparsity training scheme allows SSA to maintain efficiency and accuracy across diverse input workloads (P::\texttt{SVH} and P::\texttt{MCT}), addressing the limitations of existing spatial feature extraction modules under sensor variation~\cite{du2023saits,lai2023lightcts,wang2023networked,lai2024lightcts}.

\subsection{Output and Training of \textnormal{\AdaCTSi{}}}
\label{ssec:output}
In the final step, a multilayer perceptron (MLP) transforms the $\mathtt{N} \times \mathtt{D}$ features produced by SSA into an imputed output of shape $\mathtt{N} \times \mathtt{T}$. Each sensor's features are processed independently.
\begin{equation}
    \hat{\mathbf{X}} = \text{MLP}(\mathbf{X}_{SSA}),
\end{equation}
where $\mathbf{X}_{SSA} \in \mathbb{R}^{\mathtt{N} \times \mathtt{D}}$ is the output from the SSA module, and $\hat{\mathbf{X}} \in \mathbb{R}^{\mathtt{N} \times \mathtt{T}}$ is the final imputed output.

During training, the SSA module uses a randomly selected number of sensors for sparse spatial attention, ranging from $1$ to $\mathtt{N}$ (P::\texttt{SVH}). \AdaCTSi{} uses the Masked Mean Absolute Error (Masked MAE)~\cite{wang2023networked,du2023saits,cao2018brits} as the loss function. This metric focuses on errors in the masked (missing) regions to optimize the model:
\begin{equation}
    \text{Masked MAE} = \frac{\sum_{i=1}^{\mathtt{N}} \sum_{j=1}^{\mathtt{T}} (1 - \mathbf{M}_{i, j}) \cdot \left| \mathbf{X}_{i, j} - \hat{\mathbf{X}}_{i, j} \right|}{\sum_{i=1}^{\mathtt{N}} \sum_{j=1}^{\mathtt{T}} (1 - \mathbf{M}_{i, j})},
\end{equation}
where $\mathbf{X}_{i, j}$ and $\hat{\mathbf{X}}_{i, j}$ represent the ground-truth and imputed values, respectively, at the $i$-th sensor and the $j$-th timestamp.

\subsection{Correlation-Weighted Sensor Selection}
\label{ssec:node_selection}
To balance efficiency with the need for sufficient spatial correlation information, we introduce the Correlation-Weighted Sensor Selection (CW-SS) mechanism. When only a few sensors have missing values, those sensors alone may provide insufficient spatial context for accurate imputation. Conversely, using all available sensors can lead to excessive computational overhead. CW-SS addresses this challenge by ensuring that a minimum number of sensors, denoted by $\mathtt{K}$, are used for on-demand imputation. The parameter study of $\mathtt{K}$ is presented in \Cref{ssec:parameter}, where we provide recommendations for selecting $\mathtt{K}$. If the number of incomplete sensors falls below $\mathtt{K}$, CW-SS selectively incorporates additional sensors based on their correlations with the incomplete ones (P::\texttt{ISS}). As a result, CW-SS keeps imputation efficient and accurate even with a limited input sensor set (P::\texttt{SVH}).

\subsubsection{Correlation Calculation}
The correlations between sensors are determined using the Sensor Identity Index Table $\mathbf{IT}^S$ described in \Cref{subsubsec:LTSIT}. In this index table, each row corresponds to the embedding of a specific sensor. The correlation matrix $\mathbf{S}$ is calculated as follows.
\begin{equation}
\mathbf{S} = \mathbf{IT}^S \times (\mathbf{IT}^S)^\top,
\end{equation}
where the element $\mathbf{S}_{i,j}$ represents the correlation between sensors $i$ and $j$. This matrix helps identify sensors that provide significant spatial information about the input sensors.

\subsubsection{Sensor Selection Strategy}

To ensure that the imputer receives at least $\mathtt{K}$ sensors for adequate spatial information, we implement the following strategies:
\begin{enumerate}[leftmargin=*]
    \item \textbf{Determine the Number of Additional Sensors Needed:}
    \begin{equation}
    N_{\text{needed}} = \mathtt{K} - N_{\text{missing}},
    \end{equation}
    where $N_{\text{missing}}$ is the number of incomplete sensors.
    
    \item \textbf{Compute Sensor Correlation:}
    For each complete sensor, compute its average correlation with the incomplete sensors:
    \begin{equation}
    \mathbf{S}_{j} = \frac{1}{N_{\text{missing}}} \sum_{i} \text{cosine\_similarity}(\mathbf{E}_j, \mathbf{E}_i),
    \end{equation}
    where $\mathbf{E}_j$ is the embedding of complete sensor $j$, and $\mathbf{E}_i$ is the embedding of incomplete sensor $i$.
    
    \item \textbf{Rank and Select Sensors:}
    Rank the complete sensors by their average correlation scores $\mathbf{S}_{j}$ and select the top $N_{\text{needed}}$ sensors.
\end{enumerate}

The effectiveness of CW-SS is supported by several considerations. By using the learned embeddings from the Sensor Identity Index Table to calculate correlations among sensors, CW-SS selects informative sensors that are spatially aligned with those containing missing data, thereby maintaining high imputation accuracy and efficiency~\cite{little2019statistical}. Ensuring that at least $\mathtt{K}$ sensors are used addresses insufficient spatial context, as theoretical work on data imputation emphasizes the importance of using enough correlated sensors for accurate reconstruction~\cite{little2019statistical}. Our experimental results in \Cref{fig:memory} demonstrate that incorporating spatial information from more sensors improves imputation accuracy. Although random sensor selection is simpler, it may include sensors that are uncorrelated with the target sensors or otherwise misleading, thereby degrading imputation accuracy~\cite{hastie2009elements}. As shown in \Cref{tab:nodeselection}, this random strategy performs worse than CW-SS.

By enforcing a minimum sensor threshold and selecting only the most spatially informative sensors, CW-SS uses spatial information efficiently. This makes \AdaCTSi{} well-suited for applications where missing rates are relatively low or missing values are concentrated within a few sensors.

\subsection{Inference Complexity Analysis}
We analyze the per-window inference complexity. Let $\mathtt{N}$ be the number of sensors, $\mathtt{T}$ the window length, $\mathtt{D}$ the embedding size, and $\mathtt{U}$ the number of sensors selected by SSA ($\mathtt{U} \leq \mathtt{N}$). The One-shot TCN with $\mathtt{L}$ temporal convolution layers and kernel size $\mathtt{k}_t$ costs $\mathcal{O}(\mathtt{N}\mathtt{T}\mathtt{D}\mathtt{L}\mathtt{k}_t)$ time and $\mathcal{O}(\mathtt{N}\mathtt{T}\mathtt{D})$ memory for intermediate features. LTSIT retrieval is constant-time per sensor, yielding $\mathcal{O}(\mathtt{N}\mathtt{D})$ time and $\mathcal{O}((|\mathbf{IT}^S|+|\mathbf{IT}^T|+|\mathbf{IT}^W|)\mathtt{D})$ memory for the tables. SSA computes projections in $\mathcal{O}(\mathtt{N}\mathtt{D}^2)$ and sparse attention in $\mathcal{O}(\mathtt{U}\mathtt{N}\mathtt{D})$, reducing the standard $\mathcal{O}(\mathtt{N}^2\mathtt{D})$ cost to linear in $\mathtt{U}$. The output MLP applied per sensor costs $\mathcal{O}(\mathtt{N}\mathtt{D}\mathtt{T})$. CW-SS computes correlations between incomplete and candidate sensors in $\mathcal{O}(\mathtt{N}\mathtt{D})$ and selects the top sensors in $\mathcal{O}(\mathtt{N}\log \mathtt{N})$ (or $\mathcal{O}(\mathtt{N})$ with partial selection). Overall, inference runtime and memory scale linearly with $\mathtt{U}$ and the effective input size; when inputs are partitioned into $g$ groups, the per-pass inference cost scales with $\mathtt{N}/g$, enabling on-demand complexity tuning under resource constraints. Training follows the same asymptotic order with a constant-factor overhead due to backpropagation.

\section{Experiments}
\label{sec:experiments}

This section presents the experimental evaluation of the proposed method. \Cref{ssec:setup} describes the experimental setup, \Cref{subsec:overall} compares overall performance, and \Cref{subsec:sfada,subsec:eiada,subsec:drada} empirically validate S::\texttt{SFA}, S::\texttt{EIA}, and S::\texttt{DRA}, respectively. \Cref{subsec:forecasting} examines downstream forecasting performance; \Cref{ssec:masking_rate,ssec:efficiency} study missing-rate robustness and efficiency, respectively; \Cref{ssec:ablation} presents ablation studies; and \Cref{ssec:parameter} analyzes key parameters and provides selection recommendations. Collectively, these experiments validate the accuracy, adaptability, robustness, and efficiency of \AdaCTSi{}.

\begin{table*}[ht]
\centering
\caption{Overview of Baseline Methods.}
\resizebox{0.95\textwidth}{!}{%
\begin{tabular}{llp{0.75\textwidth}}
\toprule
\textbf{Category} & \textbf{Method} & \textbf{Description} \\ 
\midrule
\multirow{5}{*}{Traditional} 
    & \MEAN{} & Imputes missing values in a data chunk using the mean of each dimension. \\  
    & \MICE{}~\cite{white2011multiple} & Imputes missing values using multiple imputation by chained equations. \\  
    & \MF{} & Treats a data chunk as a matrix and uses iterative singular value decomposition to impute missing values. \\  
    & \TRMF{}~\cite{yu2016temporal} & Factorizes high-dimensional time series with temporal autoregressive regularization on latent factors. \\
    & \SMVNMF{}~\cite{gong2020spatial} & Applies spatial multi-view NMF to impute missing values. \\
\midrule
\multirow{2}{*}{Generative} 
    & \rGAIN{}~\cite{miao2021generative} & Uses a GAN with bidirectional encoder and decoder networks to generate missing values. \\ 
    & \PoGeVon{}~\cite{wang2023networked} & Employs a VAE with a decoder integrating GRU, self-attention, and MPNNs to impute missing values. \\
\midrule
\multirow{4}{*}{Time-Series-Based}
    & \BRITS{}~\cite{cao2018brits} & Uses bidirectional RNNs with masking to iteratively impute missing values. \\  
    & \SAITS{}~\cite{du2023saits} & Leverages self-attention to extract spatio-temporal patterns for CTS imputation. \\ 
    & \TimesNet{}~\cite{wu2022timesnet} & Transforms 1D time series into 2D tensors to capture multi-period temporal variations using CNNs. \\ 
    & \GRIN{}~\cite{andrea2022filling} & Integrates MPNNs for spatial pattern identification and RNNs for temporal patterns. \\  
\midrule
Graph-based         
    & \NET{}~\cite{jing2021network} & Employs GCNs for spatial dependencies and RNNs for temporal dependencies. \\ 
\bottomrule
\end{tabular}
}
\label{tab:baseline}
\end{table*}

\subsection{Experimental Setup}
\label{ssec:setup}

Our implementation uses PyTorch, with model architectures, datasets, and training protocols available in our open-source codebase~\cite{AdaCTSi_codebase}. Experiments were conducted on a high-performance computing system with an NVIDIA Quadro RTX 8000 GPU and an Intel Xeon Gold 5215 CPU at 2.50~GHz. Unless otherwise specified, we use the Adam optimizer (learning rate $1\times10^{-3}$, weight decay 0), a batch size of 32, gradient clipping at 5, cosine annealing with $\eta_{\min}=1\times10^{-4}$, and early stopping with patience 80. We train for 400 epochs on PeMS and 300 epochs on Beijing-AQI and Vessel. All reported results are averaged over five runs.

\noindent\textbf{Datasets}. We use five real-world CTS datasets spanning traffic, air-quality monitoring, and vessel trajectory domains. The traffic datasets are collected from the Caltrans Performance Measurement System~\cite{chen2001freeway}. PeMS-BA, PeMS-LA, and PeMS-SD correspond to the Bay Area, Los Angeles, and San Diego regional subnetworks, respectively, with records sampled every 5 minutes. The raw regional files contain 1,632, 2,383, and 674 candidate sensors; following prior CTS imputation protocols~\cite{wang2023networked, rectsi2024resource}, each experiment uses 64 selected sensors and applies a 25\% point-missing evaluation mask unless otherwise specified.

\textit{Beijing-AQI} contains hourly PM2.5 readings from 12 monitoring stations in Beijing from 2013 to 2017~\cite{zhang2017airquality}. Unlike PeMS, its evaluation mask is generated by temporal inference, where values observed in the current year but missing in an adjacent year are used as evaluation targets. This yields an evaluation missing rate of approximately 2\%. \textit{Vessel} contains AIS measurements from 20 vessels sampled every 10 seconds~\cite{grgicevic2023ais}. It has a natural missing rate of 17.49\%, and we additionally apply a sparse random evaluation mask over approximately 0.4\% of the observed points.

\begin{table}[t]
\centering
\scriptsize
\caption{Key Statistical Properties of Datasets.}
\label{tab:dataset_properties}
\resizebox{\linewidth}{!}{%
\begin{tabular}{@{}lrcccccc@{}}
\toprule
Dataset & Nodes & Mean$\pm$Std & CV & $|\rho|$ mean/p90 & Density & Avg deg. & Clust. \\
\midrule
PeMS-BA & 64 & 394.991$\pm$225.045 & 0.570 & 0.820/0.922 & 0.078 & 4.906 & 0.073 \\
PeMS-LA & 64 & 517.405$\pm$273.159 & 0.528 & 0.840/0.940 & 0.065 & 4.094 & 0.119 \\
PeMS-SD & 64 & 374.854$\pm$239.430 & 0.639 & 0.871/0.971 & 0.114 & 7.156 & 0.031 \\
Beijing-AQI & 12 & 79.793$\pm$80.822 & 1.013 & 0.886/0.957 & 0.470 & 5.167 & 0.646 \\
Vessel & 20 & 62.279$\pm$0.314 & 0.005 & 0.830/0.993 & 0.916 & 17.400 & 0.986 \\
\bottomrule
\end{tabular}%
}
\end{table}

\noindent\textbf{Dataset Characterization}. \Cref{tab:dataset_properties} reports complementary statistical and topological properties that shape imputation difficulty. Graph statistics use the provided adjacency when available and otherwise a correlation graph. Within PeMS, the table highlights differences in traffic scale, variability, correlation distribution, and topology: PeMS-LA has the highest mean flow, PeMS-SD has the largest coefficient of variation and densest selected-sensor graph, and PeMS-LA and PeMS-BA have sparser selected-sensor graphs than PeMS-SD. Beijing-AQI combines high relative variability with dense station correlations, whereas Vessel has extremely low coefficient of variation and high local smoothness. These contrasts provide a cross-domain evaluation setting where imputers must handle different value scales, correlation strengths, and graph structures.

\noindent\textbf{Metrics}. We evaluate accuracy using Mean Absolute Error (MAE), Mean Squared Error (MSE), and Mean Relative Error (MRE). MAE measures overall accuracy, MSE emphasizes larger deviations, and MRE assesses relative accuracy, especially when ground-truth values are low. Lower values indicate better imputation performance.

\noindent\textbf{Baselines.} We include twelve methods for overall comparison, chosen to represent a broad range of approaches. Specifically, \MEAN{}, \MF{}, \MICE{}~\cite{white2011multiple}, \TRMF{}~\cite{yu2016temporal}, and \SMVNMF{}~\cite{gong2020spatial} exemplify classical non-DL approaches, including temporally regularized and spatial multi-view matrix factorization. \rGAIN{}~\cite{miao2021generative} and \PoGeVon{}~\cite{wang2023networked} use generative techniques such as GANs and VAEs. Models such as \BRITS{}~\cite{cao2018brits}, \TimesNet{}~\cite{wu2022timesnet}, \GRIN{}~\cite{andrea2022filling}, \NET{}~\cite{jing2021network}, and \SAITS{}~\cite{du2023saits} integrate layered temporal and spatial feature extraction modules for imputation. \Cref{tab:baseline} provides detailed descriptions of the baselines.

Together, these baselines cover classical statistical and matrix-factorization methods as well as representative recurrent, convolutional, attention-based, graph-based, and generative imputers. We evaluate them as fixed-input imputers under their standard protocols.

\subsection{Overall Performance Comparisons}
\label{subsec:overall}

To comprehensively evaluate the performance of \AdaCTSi{}, we conduct extensive comparisons against representative methods on five benchmark datasets spanning traffic, air quality, and trajectory domains.

\begin{table*}[ht]
  \centering
  \caption{Overall Accuracy Comparison (Part I).}
  \resizebox{\linewidth}{!}{
  \begin{threeparttable}
  \begin{tabular}{@{}lcccccccccc@{}}
    \toprule
    \multirow{2}{*}{Model} & \multicolumn{3}{c}{PeMS-BA} & \multicolumn{3}{c}{PeMS-LA} & \multicolumn{3}{c@{}}{PeMS-SD} \\
    \cmidrule(r){2-4} \cmidrule(r){5-7} \cmidrule(lr){8-10}
    & {MAE} & {MSE} & {MRE} & {MAE} & {MSE} & {MRE} & {MAE} & {MSE} & {MRE} \\    
    \midrule
    {\MEAN{}}      & {192.047 $\pm$ 0.000} & {47504.159 $\pm$ 0.000} & {0.474 $\pm$ 0.000} & {216.681 $\pm$ 0.000} & {62664.657 $\pm$ 0.000} & {0.406 $\pm$ 0.000} & {208.192 $\pm$ 0.000} & {55780.002 $\pm$ 0.000} & {0.529 $\pm$ 0.000} \\
    {\MF{}}        & {57.265 $\pm$ 1.148}  & {8091.407 $\pm$ 185.123}  & {0.141 $\pm$ 0.003} & {77.339 $\pm$ 0.699}  & {15202.678 $\pm$ 156.348} & {0.145 $\pm$ 0.001} & {45.811 $\pm$ 0.318}  & {6044.345 $\pm$ 72.976}  & {0.117 $\pm$ 0.001} \\ 
    {\MICE{}}      & {50.861 $\pm$ 0.765}  & {6724.148 $\pm$ 109.829}  & {0.126 $\pm$ 0.002} & {64.018 $\pm$ 1.015}  & {10822.355 $\pm$ 405.410} & {0.120 $\pm$ 0.002} & {38.978 $\pm$ 1.036}  & {4771.186 $\pm$ 92.335}  & {0.100 $\pm$ 0.003} \\ 
    {\TRMF{}}      & {71.259 $\pm$ 0.331}  & {10281.163 $\pm$ 56.114}  & {0.176 $\pm$ 0.001} & {79.180 $\pm$ 0.301}  & {13569.843 $\pm$ 60.300} & {0.149 $\pm$ 0.001} & {61.226 $\pm$ 0.252}  & {8316.444 $\pm$ 36.472}  & {0.156 $\pm$ 0.001} \\
    {\SMVNMF{}}    & {48.347 $\pm$ 0.550}  & {5874.016 $\pm$ 143.257}  & {0.120 $\pm$ 0.001} & {59.412 $\pm$ 0.550}  & {9647.162 $\pm$ 281.867} & {0.112 $\pm$ 0.001} & {38.378 $\pm$ 0.591}  & {4616.382 $\pm$ 72.088}  & {0.098 $\pm$ 0.002} \\
    \midrule
    \BRITS{}   & 30.274 $\pm$ 0.095  & 2942.411 $\pm$ 16.511  & 0.075 $\pm$ 0.000 & 36.921 $\pm$ 0.133  & 3681.595 $\pm$ 21.635  & 0.069 $\pm$ 0.000 & 21.232 $\pm$ 0.059  & 1563.234 $\pm$ 28.309  & 0.054 $\pm$ 0.000\\
    \rGAIN{}   & 38.862 $\pm$ 0.752  & 3422.914 $\pm$ 61.281  & 0.096 $\pm$ 0.002 & 49.611 $\pm$ 1.083  & 5533.964 $\pm$ 234.335  & 0.093 $\pm$ 0.002 & 33.212 $\pm$ 1.475  & 2341.466 $\pm$ 98.314  & 0.085 $\pm$ 0.004\\
    \SAITS{}   & 46.567 $\pm$ 0.530  & 5412.574 $\pm$ 161.132  & 0.115 $\pm$ 0.001 & 61.896 $\pm$ 0.892  & 10998.854 $\pm$ 204.345 & 0.116 $\pm$ 0.002 & 34.117 $\pm$ 0.886  & 4101.397 $\pm$ 152.141  & 0.087 $\pm$ 0.002\\
    \TimesNet{}& 25.859 $\pm$ 0.115  & 1676.843 $\pm$ 16.144  & 0.064 $\pm$ 0.000 & 27.452 $\pm$ 0.114  & 2058.227 $\pm$ 6.213  & 0.052 $\pm$ 0.000 & 21.583 $\pm$ 0.085  & 1284.300 $\pm$ 21.839  & 0.055 $\pm$ 0.000\\
    \GRIN{}    & 30.057 $\pm$ 1.073  & 1922.072 $\pm$ 74.327  & 0.074 $\pm$ 0.003 & 47.835 $\pm$ 2.059  & 4561.512 $\pm$ 298.533  & 0.090 $\pm$ 0.004 & 41.001 $\pm$ 1.543  & 3000.012 $\pm$ 201.018  & 0.105 $\pm$ 0.004\\
    \NET{} & 35.671 $\pm$ 0.111  & 2735.574 $\pm$ 6.138  & 0.089 $\pm$ 0.000 & 37.652 $\pm$ 0.113  & 3416.784 $\pm$ 6.765  & 0.071 $\pm$ 0.000 & 34.111 $\pm$ 0.184  & 2487.581 $\pm$ 9.798  & 0.087 $\pm$ 0.000\\
    \PoGeVon{} & \underline{22.194 $\pm$ 0.046}  & \underline{1248.681 $\pm$ 4.297}  & \underline{0.055 $\pm$ 0.000} & \underline{23.905 $\pm$ 0.245}  & \underline{1714.962 $\pm$ 31.035}  &\underline{0.045 $\pm$ 0.000} & \underline{18.990 $\pm$ 0.112}  & \underline{951.559 $\pm$ 8.264}   & \underline{0.048 $\pm$ 0.000}\\
    \midrule
    \AdaCTSi{}  & \textbf{19.433 $\pm$ 0.062}  & \textbf{1056.990 $\pm$ 5.843}  & \textbf{0.048 $\pm$ 0.000} & \textbf{22.141 $\pm$ 0.178}  & \textbf{1654.584 $\pm$ 18.762}  & \textbf{0.041 $\pm$ 0.000} & \textbf{15.945 $\pm$ 0.095}  & \textbf{754.236 $\pm$ 6.127}   & \textbf{0.041 $\pm$ 0.000}\\
    \bottomrule
  \end{tabular}
  \end{threeparttable}
   }
    \label{tab:overall_comparisons}
\end{table*}

\begin{table}[ht]
  \centering
  \caption{Overall Accuracy Comparison (Part II).}
  \resizebox{\linewidth}{!}{
  \begin{threeparttable}
  \begin{tabular}{@{}lcccccccc@{}}
    \toprule
    \multirow{2}{*}{Model} & \multicolumn{3}{c}{Beijing-AQI} & \multicolumn{3}{c}{Vessel} \\
    \cmidrule(r){2-4} \cmidrule(l){5-7}
    & {MAE} & {MSE} & {MRE} & {MAE} & {MSE} & {MRE} \\    
    \midrule
    {\MEAN{}}      & {63.953 $\pm$ 0.000} & {7670.420 $\pm$ 0.000} & {0.813 $\pm$ 0.000} & {0.129 $\pm$ 0.000} & {0.025 $\pm$ 0.000} & {0.002 $\pm$ 0.000} \\
    {\MF{}}        & {22.897 $\pm$ 0.324} & {1410.897 $\pm$ 77.050} & {0.291 $\pm$ 0.004} & {0.027 $\pm$ 0.001} & \underline{{0.001 $\pm$ 0.000}} & \underline{{0.0004 $\pm$ 0.000}} \\ 
    {\MICE{}}      & {29.779 $\pm$ 1.022} & {1997.575 $\pm$ 233.637} & {0.378 $\pm$ 0.013} & {6.391 $\pm$ 0.291} & {95.776 $\pm$ 9.147} & {0.103 $\pm$ 0.005} \\ 
    {\TRMF{}}      & {60.726 $\pm$ 0.015} & {6833.239 $\pm$ 3.863} & {0.788 $\pm$ 0.000} & {0.161 $\pm$ 0.003} & {0.035 $\pm$ 0.001} & {0.003 $\pm$ 0.000} \\
    {\SMVNMF{}}    & {14.691 $\pm$ 0.838} & {622.225 $\pm$ 89.013} & {0.191 $\pm$ 0.011} & \underline{{0.026 $\pm$ 0.002}} & {0.001 $\pm$ 0.000} & {0.0004 $\pm$ 0.000} \\
    \midrule
    \BRITS{}    & \underline{14.157 $\pm$ 0.128} & \underline{545.447 $\pm$ 12.631} & \underline{0.180 $\pm$ 0.002} & 0.079 $\pm$ 0.003 & 0.010 $\pm$ 0.001 & 0.001 $\pm$ 0.000 \\
    \rGAIN{}     & 68.960 $\pm$ 2.145 & 12230.704 $\pm$ 412.550 & 0.876 $\pm$ 0.015 & 0.203 $\pm$ 0.012 & 0.062 $\pm$ 0.005 & 0.003 $\pm$ 0.000 \\
    \SAITS{}    & 72.418 $\pm$ 1.872 & 13540.215 $\pm$ 380.440 & 0.912 $\pm$ 0.012 & 0.221 $\pm$ 0.015 & 0.075 $\pm$ 0.006 & 0.004 $\pm$ 0.000 \\
    \TimesNet{} & 28.144 $\pm$ 0.452 & 2516.321 $\pm$ 52.180 & 0.354 $\pm$ 0.006 & 0.125 $\pm$ 0.006 & 0.028 $\pm$ 0.002 & 0.002 $\pm$ 0.000 \\
    \GRIN{}     & 61.435 $\pm$ 1.055 & 7664.824 $\pm$ 185.420 & 0.781 $\pm$ 0.009 & 0.137 $\pm$ 0.008 & 0.032 $\pm$ 0.003 & 0.002 $\pm$ 0.000 \\
    \NET{}     & 61.064 $\pm$ 0.981 & 7783.898 $\pm$ 160.330 & 0.776 $\pm$ 0.008 & {0.119 $\pm$ 0.004} & {0.025 $\pm$ 0.002} & {0.002 $\pm$ 0.000} \\
    \PoGeVon{}  & 25.765 $\pm$ 0.312 & 2295.291 $\pm$ 45.882 & 0.327 $\pm$ 0.005 & 0.139 $\pm$ 0.007 & 0.035 $\pm$ 0.002 & 0.002 $\pm$ 0.000 \\
    \midrule
    \AdaCTSi{}  & \textbf{6.159 $\pm$ 0.082} & \textbf{95.818 $\pm$ 2.451} & \textbf{0.080 $\pm$ 0.001} & \textbf{0.007 $\pm$ 0.000} & \textbf{0.0001 $\pm$ 0.000} & \textbf{0.0001 $\pm$ 0.000} \\
    \bottomrule
  \end{tabular}
  \end{threeparttable}
   }
    \label{tab:overall_results_final}
\end{table}

\Cref{tab:overall_comparisons,tab:overall_results_final} demonstrate that \AdaCTSi{} consistently outperforms all baselines across the five datasets, including the matrix-factorization baselines \TRMF{} and \SMVNMF{}. On the PeMS traffic datasets, which exhibit strong spatio-temporal correlations, \AdaCTSi{} achieves an average MAE improvement of approximately 12.0\% over \PoGeVon{} (e.g., 15.945 vs. 18.990 on PeMS-SD). The Beijing-AQI and Vessel datasets further illustrate how different value scales, correlation structures, and missing patterns affect imputation accuracy.

On Beijing-AQI, \BRITS{} and \SMVNMF{} are the strongest non-\AdaCTSi{} baselines in the deep and classical groups, respectively. Several deep models show large MRE values under temporal inference, indicating limited cross-year transfer in this setting. \AdaCTSi{} achieves the best results (MAE = 6.159, a 56.5\% reduction relative to \BRITS{}) through its decoupled LTSIT design: the Sensor Identity Index Table ($\mathbf{IT}^S$) captures stable station-specific characteristics, while the Periodic Temporal Index Table ($\mathbf{IT}^T$) independently models recurring temporal patterns.

The Vessel dataset has very low relative variation, strong lag-1 smoothness, short sequences, and an extremely sparse 0.4\% evaluation mask. These properties make simple interpolation and low-rank reconstruction well matched to the data, as reflected by the strong results of \SMVNMF{} and \MF{} (MAE = 0.026 $\pm$ 0.002 and 0.027 $\pm$ 0.001, respectively). \AdaCTSi{} achieves a further improvement (MAE = 0.007) through dedicated per-vessel embeddings in $\mathbf{IT}^S$ and SSA's adaptive filtering of less informative cross-series interactions, combining sensor-specific modeling with deep temporal representation learning.

These results validate that \AdaCTSi{}'s decoupled sensor-specific embedding design provides essential flexibility across diverse spatial correlation structures and missing patterns.
\subsection{Empirical Verification for S::\texttt{SFA}}
\label{subsec:sfada}

In real-world sensor networks, sensor failures leave some nodes unobserved and reduce the spatial context available for imputation. To assess sensor-failure adaptability (S::\texttt{SFA}), we conduct a PeMS-SD stress test by varying the failed-sensor ratio from 12.5\% to 50\%. All methods are evaluated under the same failed-sensor sets.

We compare \AdaCTSi{} with three representative CTS imputers: \PoGeVon{}, \TimesNet{}, and \BRITS{}, covering generative/self-attention, convolutional, and recurrent designs. For the fixed-input baselines, failed sensors are represented by zeroed values and missing masks, and metric computation omits outputs for failed sensors.

\begin{table}[ht]
\centering
\caption{Sensor-Failure Adaptability Study on PeMS-SD.}
\resizebox{\linewidth}{!}{
\begin{threeparttable}
\begin{tabular}{@{}lcccccccccc@{}}
\toprule
\multirow{2}{*}{Model} & \multicolumn{3}{c}{12.5\% (8/64) Failed} & \multicolumn{3}{c}{25\% (16/64) Failed} & \multicolumn{3}{c}{50\% (32/64) Failed} \\
\cmidrule(r){2-4} \cmidrule(r){5-7} \cmidrule(l){8-10}
& {MAE} & {MSE} & {MRE} & {MAE} & {MSE} & {MRE} & {MAE} & {MSE} & {MRE} \\
\midrule
\BRITS{} & 22.770 & 1430.077 & 0.057 & 22.868 & 1443.129 & 0.062 & 23.910 & 1532.279 & 0.066\\
\TimesNet{} & 43.377 & 4889.382 & 0.111 & 54.791 & 7017.141 & 0.140 & 106.984 & 17557.797 & 0.273\\
\PoGeVon{} & \underline{19.346} & \underline{779.665} & \underline{0.051} & \underline{19.911} & \underline{824.927} & \underline{0.052} & \underline{22.650} & \underline{1021.497} & \underline{0.061}\\
\midrule
\AdaCTSi{} & \textbf{16.525} & \textbf{611.174} & \textbf{0.043} & \textbf{16.899} & \textbf{638.036} & \textbf{0.044} & \textbf{17.286} & \textbf{669.428} & \textbf{0.046}\\
\bottomrule
\end{tabular}
\end{threeparttable}
}
\label{tab:drift}
\end{table}

\Cref{tab:drift} shows that the errors of all imputers increase with the failed-sensor ratio because less spatial information is available. \AdaCTSi{} consistently achieves the lowest MAE, MSE, and MRE across all ratios. \PoGeVon{} remains the strongest baseline, \BRITS{} is relatively stable, and \TimesNet{} degrades more sharply as failures increase. These results show that \AdaCTSi{} preserves strong imputation accuracy under S::\texttt{{SFA}}.

\subsection{Empirical Verification for S::\texttt{{EIA}}}
\label{subsec:eiada}
In practical applications, only a small subset of sensors may require imputation at a given time. Full-sensor processing can therefore introduce unnecessary computation. CW-SS addresses this selective-imputation setting by augmenting the incomplete sensors with spatially informative complete sensors only when the incomplete set is smaller than a threshold $\mathtt{K}$.

We isolate the contribution of CW-SS under a 2\% evaluation missing rate. Missing-rate sensitivity is studied separately in \Cref{ssec:masking_rate}. We compare four \AdaCTSi{} variants: \AdaCTSionly{} uses only incomplete sensors; \AdaCTSirand{} adds random complete sensors up to $\mathtt{K}$; \AdaCTSicw{} selects additional sensors by CW-SS; and \AdaCTSiall{} uses all sensors as a full-sensor reference. We set $\mathtt{K}=32$ for the selective variants. The table also lists \PoGeVon{} under its standard full-sensor setting.

\begin{table}[ht]
  \centering
  \caption{CW-SS Sensor-Selection Ablation.}

    \resizebox{\linewidth}{!}{
  \begin{threeparttable}
  \begin{tabular}{@{}lcccccccccc@{}}
    \toprule
    \multirow{2}{*}{Model} & \multicolumn{3}{c}{PeMS-BA} & \multicolumn{3}{c}{PeMS-LA} & \multicolumn{3}{c@{}}{PeMS-SD} \\
    \cmidrule(r){2-4} \cmidrule(r){5-7} \cmidrule(lr){8-10}
    & {MAE} & {MSE} & {MRE} & {MAE} & {MSE} & {MRE} & {MAE} & {MSE} & {MRE} \\    
    \midrule
    \PoGeVon{} & {20.838} & {1177.320} & {0.052} & {22.706} & {1510.112} & {0.043} & {17.112} & {742.957} & {0.044}\\
    \AdaCTSiall{}  & {17.384}  & {941.174}  & {0.043} & {20.626}  & {1389.273}  & {0.039} & {14.016}  & {564.885}   & {0.036}\\
        \midrule

    \AdaCTSionly{} & {20.717}  & {1181.736}  & {0.052} & {23.293}  & {1601.004}  & {0.044} & {17.524}  & {783.050}   & {0.045}\\
    \AdaCTSirand{} & \underline{20.043} & \underline{1121.233} & \underline{0.050} & \underline{22.229} & \underline{1471.157} & \textbf{0.042} & \underline{16.654} & \underline{716.679} & \underline{0.043}\\
    \AdaCTSicw{} & \textbf{19.512} & \textbf{1081.589} & \textbf{0.049} & \textbf{22.163} & \textbf{1470.742} & \textbf{0.042} & \textbf{16.376} & \textbf{693.495} & \textbf{0.042}\\
    \bottomrule
  \end{tabular}
  \end{threeparttable}
   }
    \label{tab:nodeselection}
\end{table}

\Cref{tab:nodeselection} shows that \AdaCTSiall{} achieves the lowest error with full-sensor input. Among the selective variants, \AdaCTSicw{} performs best on all three PeMS datasets, outperforming both \AdaCTSionly{} and \AdaCTSirand{}. It also attains lower MAE than \PoGeVon{} under full-sensor input, showing the value of selecting informative spatial context for selective imputation. This confirms the specific contribution of CW-SS to S::\texttt{{EIA}}.

\subsection{Empirical Verification for S::\texttt{{DRA}}}
\label{subsec:drada}
In real-world applications, imputers must operate across devices with varying memory and processing capacities. To evaluate how \AdaCTSi{} addresses such computational mismatches on demand, we focus on three aspects: adaptation to memory capacity, adaptation to processing power, and the balance between input workload and model complexity.

\subsubsection{Adaptation to Memory Capacity}
\begin{figure*}[ht]
	\centering
	\includegraphics[width=\linewidth]{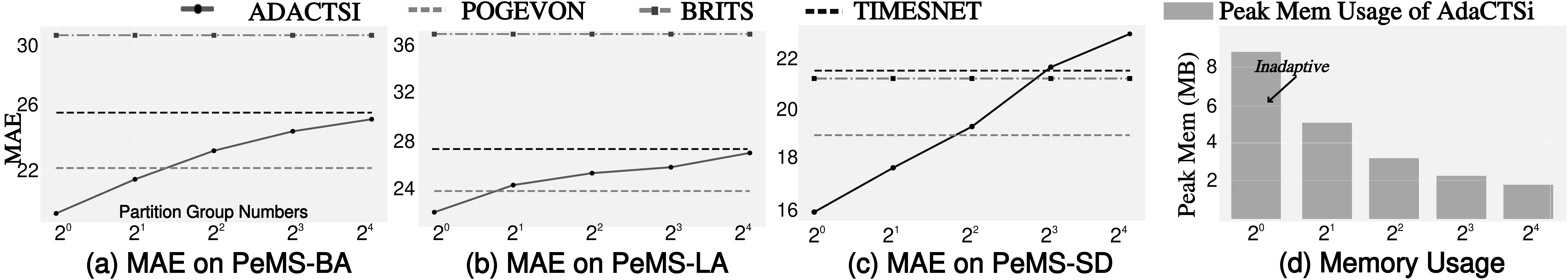}
        \caption{MAE and peak memory usage of \AdaCTSi{} with different numbers of sensor-partition groups ($g=1, 2, 4, 8, 16$). Horizontal baseline lines are fixed full-input references from \Cref{tab:overall_comparisons}.}

 \label{fig:memory}
\end{figure*}

Devices such as smart meters, wearables, and underwater monitors often have limited and varying memory capacities. To assess how \AdaCTSi{} adapts to different memory resources on demand, we simulate memory constraints by partitioning the input CTS data along the sensor dimension. Specifically, we divide the input window into $g = 1, 2, 4, 8, 16$ groups, each containing an equal number of sensors. These partitions are sequentially fed into the trained \AdaCTSi{} model, simulating devices with different memory capacities. As illustrated in \Cref{fig:memory}~(a)--(c), increasing the number of partition groups slightly degrades accuracy but keeps \AdaCTSi{} competitive with strong full-input baselines. We include the fixed full-input accuracies of \PoGeVon{}, \BRITS{}, and \TimesNet{} from \Cref{tab:overall_comparisons} as horizontal references for this sensor-partition trade-off.

As illustrated in \Cref{fig:memory}~(d), increasing the number of groups reduces the per-pass sensor-partition size and lowers peak memory usage. This partitioning is a resource-control mechanism rather than an accuracy-enhancement step: smaller partitions reduce memory but slightly increase MAE because spatial correlations can be captured only within each processed sensor group. Nevertheless, performance remains competitive with baseline methods. Notably, when $g = 16$, memory demand falls below 2 MB, enabling \AdaCTSi{} to be deployed on MCUs such as the STM32H7~\cite{lai2024e2usd,lai2023lightcts,STMCU}. This result demonstrates a favorable trade-off between memory usage and accuracy. The ability to adjust memory demand enhances \AdaCTSi{}'s usability on resource-constrained devices and broadens its application range.

\subsubsection{Adaptation to Processing Power}
To evaluate the adaptability of \AdaCTSi{} to different levels of processing power, we adjust the attention sparsity, which determines the number of sensors used in sparse spatial attention and directly influences computational demand. By setting the attention sparsity rate to 0\%, 25\%, 50\%, or 75\%, we simulate environments with different processing capacities. A higher sparsity rate excludes a larger proportion of sensors from self-attention, thereby reducing computational demand.

\begin{table}[ht]
\centering
\caption{Accuracy at Different Attention Sparsity Rates.}
\resizebox{\linewidth}{!}{
\begin{tabular}{@{}lccccccccc@{}}
\toprule
\multirow{2}{*}{{Sparsity}} & \multicolumn{3}{c}{{PeMS-BA}} & \multicolumn{3}{c}{{PeMS-LA}} & \multicolumn{3}{c}{{PeMS-SD}} \\
\cmidrule(lr){2-4} \cmidrule(lr){5-7} \cmidrule(lr){8-10}
& {MAE} & {MSE} & {MRE} & {MAE} & {MSE} & {MRE} & {MAE} & {MSE} & {MRE} \\
\midrule
\PoGeVon{} & 22.194 & 1248.68 & 0.055 & 23.905 & 1714.96 & 0.045 & 18.990 & 951.56 & 0.048 \\ \midrule
\AdaCTSi{}-75\% & 22.741 & 1287.47 & 0.056 & 23.109 & 1762.38 & 0.044 & 18.150 & 906.05 & 0.046 \\
\AdaCTSi{}-50\% & 20.745 & 1176.77 & 0.050 & 22.801 & 1784.14 & 0.043 & 17.950 & 889.47 & 0.046 \\
\AdaCTSi{}-25\% & \underline{19.505} & \underline{1061.09} & \textbf{0.048} & \underline{22.565} & \underline{1774.18} & \underline{0.042} & \underline{16.142} & \underline{767.30} & \textbf{0.041} \\
\AdaCTSi{}-0\% & \textbf{19.433} & \textbf{1056.99} & \textbf{0.048} & \textbf{22.143} & \textbf{1654.58} & \textbf{0.041} & \textbf{15.945} & \textbf{754.23} & \textbf{0.041} \\
\bottomrule
\end{tabular}
}
\label{tab:processing_power}
\end{table}
\Cref{tab:processing_power} demonstrates that \AdaCTSi{} maintains acceptable imputation accuracy as the attention sparsity rate increases and processing demand decreases. Using all sensors (i.e., a sparsity rate of 0\%) achieves the highest accuracy but requires spatial attention over all input sensors and thus greater processing power. At a sparsity rate of 75\%, SSA uses only 25\% of the sensors and incurs only a modest performance decrease, remaining comparable to the strongest baseline, \PoGeVon{} (e.g., MAE values of 22.741 vs. 22.194 on PeMS-BA). These findings show that \AdaCTSi{} can balance processing demand and imputation accuracy on demand. This adaptability makes \AdaCTSi{} well suited to elastic computing applications, in which processing power scales with resource availability.

\subsubsection{Balancing Input Workload with Model Complexity}
As the number of input sensors---and hence the workload---varies, adjusting model complexity is essential for optimizing performance under stable but limited resource conditions. In this experiment, we explore how \AdaCTSi{} balances input workload and model complexity on demand to maintain high accuracy while satisfying fixed resource constraints.

We vary the input workload by adjusting the sensor-partition size and modify model complexity by altering attention sparsity, following the settings used in the preceding experiments. This approach allows us to observe how different combinations of workload and complexity affect model accuracy while resource demands remain stable.

\begin{figure}[ht]
	\begin{center}
		\includegraphics[width=0.7\linewidth]{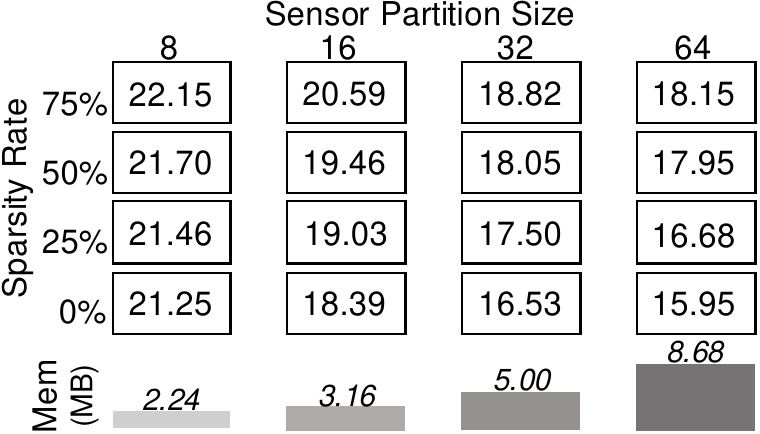}
	\end{center}
	\caption{Trade-off Between Input Workload (i.e., Sensor Partition Size) and Model Complexity (i.e., Sparsity Rate).}
	\label{fig:tradeoff}
\end{figure}

As illustrated in \Cref{fig:tradeoff}, \AdaCTSi{} balances input workload and model complexity to maintain robust on-demand performance. Each block represents a combination of input partition size (workload) and attention sparsity rate (model complexity). The primary factor determining peak memory usage is the input workload because a larger number of input sensors increases memory demand. Consequently, \AdaCTSi{} adjusts its computational demands in response to online workloads, ensuring efficient resource utilization while sustaining high imputation accuracy.

Overall, these experiments demonstrate that \AdaCTSi{} can adapt on demand to online environments with varying memory capacities, levels of processing power, and combinations of input workload and model complexity. This adaptability enables \AdaCTSi{} to operate efficiently across a wide range of devices and settings (S::\texttt{{DRA}}) while maintaining high imputation accuracy and effective resource utilization.

\subsection{Verification Using Downstream Forecasting}
\label{subsec:forecasting}

To evaluate the effectiveness of \AdaCTSi{} in improving data quality, we conduct one-step-ahead forecasting experiments. Specifically, we compare \AdaCTSi{} against the top three baselines, \PoGeVon{}, \TimesNet{}, and \BRITS{}, as identified in \Cref{subsec:overall}. We also include a No-Imputation baseline, in which missing values are filled with zeros, to evaluate the benefits of learned imputation. Each method fills missing values before forecasting, ensuring a fair and consistent comparison of its effect on forecasting accuracy.

Using the PeMS-SD dataset, we partition the data into training and testing sets at a 60:40 ratio. We explore two typical historical window sizes: a short-term window of 3 time steps (15 minutes) and a long-term window of 24 time steps (2 hours). In both cases, we forecast the next time step (5 minutes ahead). Forecasting accuracy is evaluated against the ground-truth values following established practices in CTS forecasting~\cite{lai2023lightcts,wu2021autocts}.

For forecasting, we employ widely used models, including \textsf{Random Forest (RF)}~\cite{breiman2001random}, \textsf{Gradient Boosting (GB)}~\cite{friedman2001greedy}, \textsf{XGBoost}~\cite{chen2016xgboost}, \textsf{LightGBM}~\cite{ke2017lightgbm}, and \textsf{Decision Tree (DT)}~\cite{breiman2017classification}. These models provide diverse perspectives on the quality of the imputed data.

\begin{table}[ht]
    \centering
    \caption{Downstream Forecasting Results on PeMS-SD.}
    \resizebox{\linewidth}{!}{
    \begin{tabular}{llcccccc}
        \toprule
        \multirow{2}{*}{Forecasting} & \multirow{2}{*}{Imputation} & \multicolumn{3}{c}{Historical Window Size $= 3$} & \multicolumn{3}{c}{Historical Window Size $= 24$} \\
        \cmidrule(lr){3-5} \cmidrule(lr){6-8}
        & & MAE & MSE & MAPE & MAE & MSE & MAPE \\
        \midrule
        \multirow{5}{*}{\textsf{RF}} 
        & No Imputation & 23.519 & 1100.674 & 9.36\% & 26.072 & 1358.583 & 9.87\% \\
        & \BRITS{} & 21.854 & 925.012 & \textbf{8.25\%} & 24.171 & 1115.385 & 8.88\% \\
        & \TimesNet{} & 21.207 & 838.800 & 8.69\% & 21.854 & 880.299 & 8.89\% \\
        & \PoGeVon{} & \underline{20.874} & \underline{808.613} & 8.34\% & \underline{21.011} & \underline{819.290} & \underline{8.36\%} \\
        & \AdaCTSi{} & \textbf{20.642} & \textbf{793.155} & \underline{8.29\%} & \textbf{20.597} & \textbf{786.917} & \textbf{8.22\%} \\
        \midrule
        \multirow{5}{*}{\textsf{GB}} 
        & No Imputation & 28.711 & 1560.026 & 11.69\% & 30.593 & 1824.499 & 11.96\% \\
        & \BRITS{} & 21.424 & 858.889 & \textbf{8.12\%} & 22.271 & 924.938 & 8.54\% \\
        & \TimesNet{} & 21.307 & 842.096 & 8.92\% & 21.247 & 839.423 & 8.88\% \\
        & \PoGeVon{} & \underline{21.118} & \underline{822.278} & \underline{8.59\%} & \underline{20.956} & \underline{817.093} & \underline{8.43\%} \\
        & \AdaCTSi{} & \textbf{20.696} & \textbf{791.626} & 8.63\% & \textbf{20.584} & \textbf{795.069} & \textbf{8.38\%} \\
        \midrule
        \multirow{5}{*}{\textsf{XGBoost}} 
        & No Imputation & 27.644 & 1482.383 & 11.46\% & 29.800 & 1791.796 & 11.33\% \\
        & \BRITS{} & 24.631 & 1144.539 & 9.28\% & 25.047 & 1141.652 & 9.41\% \\
        & \TimesNet{} & 22.727 & 998.016 & 9.13\% & 23.153 & 1077.440 & 9.09\% \\
        & \PoGeVon{} & \underline{22.061} & \textbf{928.763} & \underline{8.72\%} & \underline{22.783} & \underline{1015.970} & \textbf{8.78\%} \\
        & \AdaCTSi{} & \textbf{22.005} & \underline{931.100} & \textbf{8.69\%} & \textbf{22.634} & \textbf{1000.774} & \underline{8.79\%} \\
        \midrule
        \multirow{5}{*}{\textsf{LightGBM}}
        & No Imputation & 24.650 & 1180.375 & 10.26\% & 26.091 & 1358.139 & 9.90\% \\
        & \BRITS{} & 22.080 & 897.033 & 8.86\% & 23.492 & 1019.098 & 8.91\% \\
        & \TimesNet{} & 20.938 & 826.105 & 8.59\% & 20.926 & 819.107 & 8.57\% \\
        & \PoGeVon{} & \underline{20.324} & \underline{762.179} & \underline{8.15\%} & \underline{20.275} & \underline{771.485} & \underline{8.08\%} \\
        & \AdaCTSi{} & \textbf{20.146} & \textbf{752.057} & \textbf{8.11\%} & \textbf{19.637} & \textbf{709.857} & \textbf{7.96\%} \\
        \midrule
        \multirow{5}{*}{\textsf{DT}} 
        & No Imputation & 37.188 & 3298.900 & 14.79\% & 41.742 & 3803.346 & 15.50\% \\
        & \BRITS{} & 34.201 & 2172.071 & 12.68\% & 34.083 & 2439.364 & 13.20\% \\
        & \TimesNet{} & 30.776 & 1829.297 & 12.67\% & 31.921 & 1919.124 & 13.28\% \\
        & \PoGeVon{} & \underline{30.882} & \underline{1756.827} & \underline{12.53\%} & \underline{31.487} & \underline{1918.582} & \underline{12.38\%} \\
        & \AdaCTSi{} & \textbf{29.408} & \textbf{1643.930} & \textbf{11.65\%} & \textbf{29.816} & \textbf{1730.376} & \textbf{11.83\%} \\
        \bottomrule
    \end{tabular}
    }
    \label{tab:forecasting_results}
\end{table}

\Cref{tab:forecasting_results} demonstrates that \AdaCTSi{} generally outperforms the baseline methods across forecasting models and historical window sizes. In most cases, \AdaCTSi{} achieves the lowest MAE, MSE, and MAPE, with particularly notable improvements for \textsf{LightGBM} and \textsf{Random Forest}. In these cases, it surpasses leading approaches such as \PoGeVon{}, \TimesNet{}, and \BRITS{}, demonstrating its effectiveness across diverse forecasting scenarios.

A longer historical window provides more temporal information but does not consistently improve forecasting accuracy, potentially because of overfitting or the inclusion of outdated information. This observation underscores the importance of high-quality CTS imputation rather than merely increasing data volume; effective imputers such as \AdaCTSi{} support accurate forecasts across a range of models and configurations.

These results confirm that the accurate imputations produced by \AdaCTSi{} yield strong downstream forecasting performance. By providing reliable inputs, \AdaCTSi{} helps forecasting models generate better predictions, underscoring its practical value in applications with incomplete data.

\subsection{Robustness Study}
\label{ssec:masking_rate}

To evaluate the robustness of \AdaCTSi{} under different levels of data scarcity, we conduct a sensitivity analysis on PeMS-BA, PeMS-LA, and PeMS-SD using missing rates from 15\% to 85\% in 10-percentage-point increments.

\begin{figure}[t]
    \centering
    \includegraphics[width=\linewidth]{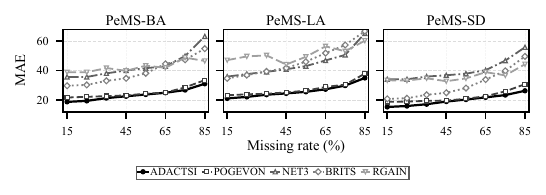}
    \caption{Impact of missing rate.}
    \label{fig:masking_rate}
\end{figure}

As illustrated in \Cref{fig:masking_rate}, the MAE of all methods generally increases as the missing rate grows, reflecting the inherent difficulty of imputation when fewer observations are available. \AdaCTSi{} consistently achieves the lowest MAE, reducing error by 7.1\% on average compared with the strongest baseline. Even at high missing rates of 55\%--85\%, it reduces MAE by 5.4\% on average. This trend demonstrates the effectiveness and robustness of our adaptive architecture under increasingly sparse observations.

\subsection{Efficiency Study} 
\label{ssec:efficiency}
We evaluate efficiency through offline training and online inference costs. \Cref{tab:training_efficiency} reports PeMS results under a common training protocol.

\begin{table}[H]
\centering
\caption{Efficiency Comparison on PeMS.}
\label{tab:training_efficiency}
\scriptsize
\setlength{\tabcolsep}{1.2pt}
\begin{tabular}{@{}lrrrr@{}}
\toprule
\multirow{2}{*}{Method} & \multicolumn{2}{c}{Training cost} & \multicolumn{2}{c@{}}{Inference cost} \\
\cmidrule(lr){2-3} \cmidrule(l){4-5}
& Time (h) & Peak Mem. (MB) & Lat. (ms) & Peak Mem. (MB) \\
\midrule
\PoGeVon{} & 39.62 & 10867.5 & 172.82 & 16.08 \\
\NET{} & 65.49 & 593.8 & 130.19 & \multicolumn{1}{c}{-} \\
\GRIN{} & 23.01 & 919.4 & 127.81 & 12.62 \\
\SAITS{} & 4.00 & 83.0 & 45.23 & 7.35 \\
\BRITS{} & 7.52 & 61.6 & 40.65 & 2.69 \\
\rGAIN{} & 3.47 & 84.2 & 18.82 & 5.02 \\
\TimesNet{} & 4.43 & 75.3 & 10.71 & 6.24 \\
\midrule
\AdaCTSi{} & 3.71 & 249.1 & 9.57 & 8.68 \\
\bottomrule
\end{tabular}
\vspace{0.3mm}
\end{table}

\Cref{tab:training_efficiency} reports training memory as peak CUDA allocation and inference memory using batch-1 \texttt{torchinfo} profiling~\cite{torchinfo}; the dash for \NET{} denotes a \texttt{torchinfo} incompatibility. \AdaCTSi{} combines adaptive sensor-subset and resource-aware inference with competitive efficiency. Its training cost is close to lightweight deep baselines and much lower than graph-heavy or generative models. For online deployment, \AdaCTSi{} achieves the lowest inference latency among the deep models while keeping a modest memory footprint. These results support the intended deployment pattern: offline training can be completed once on a server, and the resulting model can be reused across resource settings.

\subsection{Ablation Studies}
\label{ssec:ablation}
To evaluate the contribution of each component in \AdaCTSi{}, we conduct ablation studies on the three PeMS datasets. By systematically removing key components, we assess the effect of each module on imputation performance. Specifically, we create the following variants: \textbf{\AdaCTSiLTSIT{}} removes LTSIT, leaving the One-shot TCN to process the raw input without learned multi-view embeddings; \textbf{\AdaCTSiOSTCN{}} replaces the One-shot TCN with a standard MLP that lacks temporal convolutions; and \textbf{\AdaCTSiSSA{}} removes the SSA module for spatial correlation extraction.

The quantitative results in \Cref{tab:ablation} show that the complete \AdaCTSi{} architecture outperforms all ablated variants, demonstrating the contribution of each module. The higher errors of \AdaCTSiLTSIT{} confirm the role of LTSIT in learning multi-view embeddings. Similarly, the degradation of \AdaCTSiOSTCN{} demonstrates the importance of the temporal convolutions in the One-shot TCN, while the gap between \AdaCTSiSSA{} and the full model confirms the value of SSA for capturing spatial correlations among sensors.

Overall, the ablation studies validate the contribution of each component in \AdaCTSi{}. The full model outperforms all ablated variants, showing that integrating LTSIT, One-shot TCN, and SSA improves imputation performance.

\begin{table}[ht]
  \centering
  \caption{Ablation Study Results.}
  \begin{threeparttable}
     \resizebox{\linewidth}{!}{

  \begin{tabular}{@{}lccccccccc@{}}
    \toprule
    \multirow{2}{*}{Model} & \multicolumn{3}{c@{}}{PeMS-BA} & \multicolumn{3}{c@{}}{PeMS-LA} & \multicolumn{3}{c@{}}{PeMS-SD} \\
    \cmidrule(r){2-4} \cmidrule(r){5-7} \cmidrule(r){8-10}
    & {MAE} & {MSE} & {MRE} & {MAE} & {MSE} & {MRE} & {MAE} & {MSE} & {MRE} \\    
    \midrule
    \AdaCTSiLTSIT{} & 21.271  & 1124.562  & 0.053 & 26.509 & 1899.853 & 0.055 & 19.925  & 1039.318  & 0.051\\
    \AdaCTSiOSTCN{} & 26.412 & 1812.084 & 0.065 & 27.675 & 2104.829 & 0.057 & 24.518 & 1716.217 & 0.063\\
    \AdaCTSiSSA{} & 20.234 & 1081.424 & 0.050 & 23.890 & 1685.953 & 0.050 & 16.599 & 803.075 & 0.042\\
    \midrule
    \AdaCTSi{}  & \textbf{19.433}  & \textbf{1056.990}   & \textbf{0.048} & \textbf{22.141} & \textbf{1654.584} & \textbf{0.041} & \textbf{15.945}  & \textbf{754.236}   & \textbf{0.041}\\
    \bottomrule
  \end{tabular}}
  \end{threeparttable}
  \label{tab:ablation}
\end{table}

\subsection{Parameter Studies}
\label{ssec:parameter}
We investigate the influence of three key parameters---the embedding size $\mathtt{D}$, window size $\mathtt{T}$, and CW threshold $\mathtt{K}$---on imputation performance. We use MAE on the PeMS-SD dataset under the experimental configurations described in \Cref{ssec:setup,subsec:eiada}.

\begin{figure}[ht]
    \begin{center}
        \includegraphics[width=\linewidth]{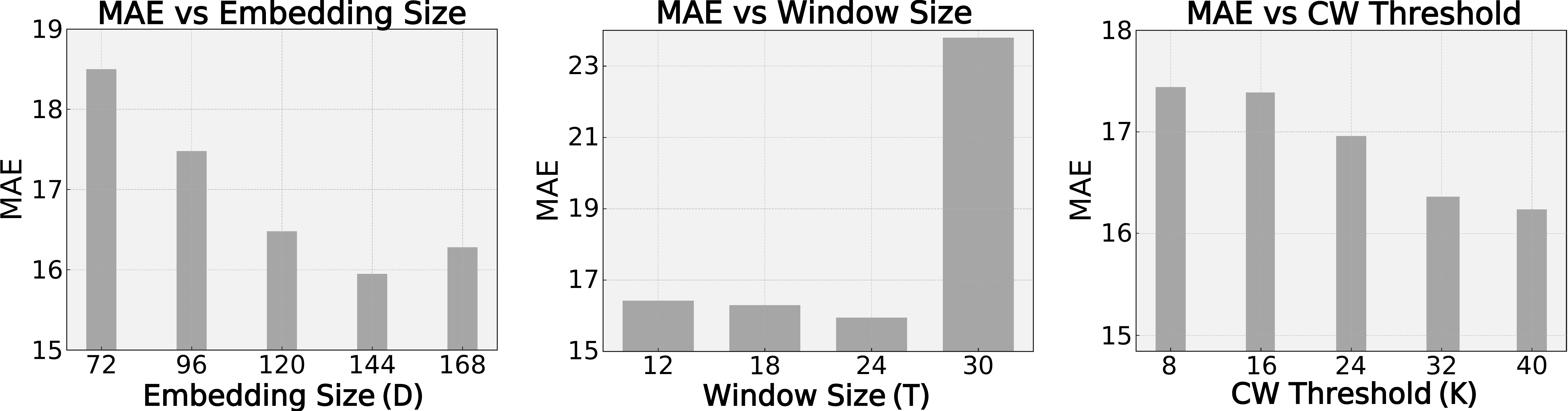}
    \end{center}
    \caption{Impact of $\mathtt{D}$, $\mathtt{T}$, and $\mathtt{K}$.}
    \label{fig:combined}
\end{figure}

\emph{Embedding Size $\mathtt{D}$}: As illustrated in \Cref{fig:combined} (left), increasing the embedding size $\mathtt{D}$ from 72 to 144 consistently reduces the MAE, reflecting improved imputation accuracy. The optimal embedding size is observed at $\mathtt{D} = 144$, which effectively balances model expressiveness and regularization, thereby minimizing the risk of overfitting. However, further increasing $\mathtt{D}$ beyond 144 results in a slight degradation in performance. This decline indicates that excessively large embeddings may introduce redundant parameters, leading to increased computational complexity, potential overfitting, and diminished generalization capability.

\emph{Window Size $\mathtt{T}$}: The impact of varying $\mathtt{T}$ is presented in \Cref{fig:combined} (middle). As the window size increases from 12 to 24, MAE decreases, indicating a benefit from the richer temporal context. The best performance is achieved at $\mathtt{T} = 24$, which represents two hours of data at five-minute sampling intervals. Window sizes of 12, 18, and 24 divide evenly into the 288-step daily cycle, whereas a window size of 30 does not and yields a higher MAE. These results suggest that both temporal context and alignment with the data's periodic structure affect model performance.

\emph{CW Threshold $\mathtt{K}$}: The sensitivity of the model to $\mathtt{K}$ is depicted in \Cref{fig:combined} (right). MAE stabilizes when $\mathtt{K} \geq 32$. Given a total of 64 sensors, setting $\mathtt{K}=32$ ensures the selection of at least half of the sensors, providing adequate spatial information for accurate imputation. Thresholds below 32 select fewer sensors, limit the available spatial context, and increase MAE. We therefore recommend setting the CW threshold to at least half of the sensor count.

Overall, \AdaCTSi{} exposes only a small set of key parameters through its streamlined architecture. One-shot TCN, LTSIT, SSA, and CW-SS are each used once rather than repeated across stacked modules. This design simplifies model configuration and deployment across diverse environments.

\section{Related Work}
\label{sec:related}

\noindent\textbf{Deep Models for CTS Imputation}.  
CTS imputation has been studied through statistical, matrix-factorization, subspace, and DL-based approaches. Representative classical baselines include \MEAN{}, Matrix Factorization (\MF{}), and Multivariate Imputation by Chained Equations (\MICE{})~\cite{white2011multiple}. Classical matrix-completion and subspace-based methods introduce structured priors through spectral regularization for low-rank matrix completion~\cite{mazumder2010spectral}, streaming pattern discovery in multivariate time series~\cite{papadimitriou2005streaming}, temporally regularized matrix factorization (TRMF)~\cite{yu2016temporal}, ST-MVL for geo-sensory time series~\cite{yi2016st}, spatial multi-view NMF (SMV-NMF) for sensor and urban statistical data~\cite{gong2020spatial}, and scalable recovery of missing blocks under different cross-correlation regimes~\cite{khayati2020scalable}. These approaches remain useful when their statistical or low-rank assumptions match the data, whereas modern large-scale CTS deployments often involve dynamic spatio-temporal correlations and heterogeneous missing patterns that benefit from representation learning.

DL models capture such complex patterns using modules such as the bidirectional RNNs in \BRITS{}~\cite{cao2018brits}, which incorporate temporal information from both past and future timestamps. More recent models enhance spatial feature extraction using GNNs~\cite{jing2021network,andrea2022filling,li2023missing} and self-attention mechanisms~\cite{du2023saits}, thereby capturing both spatial and temporal relationships~\cite{wu2022timesnet}. Generative models, including GANs~\cite{miao2021generative,luo2019e2gan} and VAEs~\cite{fortuin2020gp,wang2023networked}, learn latent representations to generate imputed values. Diffusion models~\cite{tashiro2021csdi,liu2023pristi,wang2023observed} further improve generative modeling through iterative sampling and are often studied in offline imputation settings. \RECTSI{}~\cite{rectsi2024resource} studies efficient CTS imputation under a fixed input and model configuration.

Most of these imputers are evaluated with fixed sensor sets and model configurations. In contrast, \AdaCTSi{} uses one trained model for elastic inference across variable sensor subsets and resource budgets. It combines One-shot TCN and LTSIT for sensor-wise representations, SSA for spatial correlation extraction, and CW-SS for informative-sensor selection.

\smallskip
\noindent\textbf{Adaptability in DL Methods}.  
Adaptability is essential for deploying DL models in environments with varying computational resources and input workloads. It requires balancing computational efficiency with accuracy, handling diverse workloads, and supporting devices with different resource capacities.

Existing solutions predominantly maintain a collection of models and select one based on the input workload and available resources. Although this approach is adaptable, designing, training, and updating multiple models incurs substantial computational and storage overhead. In computer vision, some methods use patch-wise processing~\cite{dosovitskiy2021an,liu2021swin,zhang2021rest}, dividing images into patches to handle varying image sizes by leveraging local spatial characteristics and texture patterns. This idea is less directly applicable to CTS imputation because CTS workloads comprise disjoint time series subsets with distinct characteristics and patterns~\cite{lai2023lightcts}.

In summary, although model collections and patch-wise processing can enhance adaptability, they introduce model-management complexity and do not directly accommodate disjoint CTS subsets. To address these issues, we propose \AdaCTSi{}, an adaptive CTS imputer for changing environments. \AdaCTSi{} features an On-demand Mode that uses a single adaptable $N$-in-1 model across varying sensor subsets and resource settings.

\section{Conclusion and Future Work}
\label{sec:conclusion}
In this paper, we presented \AdaCTSi{}, an adaptive and efficient CTS imputer for changing environments. It addresses sensor failures, excessive imputation, and resource variability by dynamically adjusting its input workload and computational requirements, thereby establishing a foundation for adaptive CTS imputation.

In future work, we will incorporate predictive resource provisioning and scheduling to anticipate workload and resource changes, enhance on-demand adaptability, and validate these improvements in real-world online scenarios. This direction involves combining workload and resource prediction with dynamic cluster adjustment, which is beyond the current scope but important for complex, real-time IoT applications. We also aim to integrate lightweight modules to improve \AdaCTSi{}'s efficiency and enable faster on-demand CTS imputation.

\section*{Acknowledgment}
This work was supported by the Fujian Provincial Artificial Intelligence Industry Development Technology Project under Grant 2025H0042 and by the Young Scientists Fund of the National Natural Science Foundation of China under Grant 62402420 for the project ``Research on Key Technologies for Edge-Intelligent Time-Series Data Imputation.''


\bibliographystyle{IEEEtran}
\bibliography{references.bib}

\makeatletter
\def\@IEEEBIOskipN{0pt}
\makeatother

\begin{IEEEbiography}[{\includegraphics[width=1in,height=1.25in,clip,keepaspectratio]{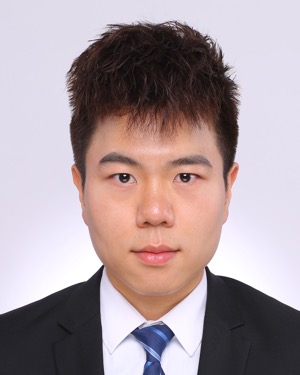}}]{Zhichen Lai}
received the bachelor's and master's degrees in Computer Science from the University of Electronic Science and Technology of China and Sichuan University, China, in 2018 and 2021, respectively, and the PhD degree in Computer Science from Aalborg University, Denmark, in 2025. He is currently an Associate Research Professor with the College of Computer and Data Science, Fuzhou University, China. In 2025, he was selected for China's Overseas Talent Recruitment Program. His research focuses on spatio-temporal data mining.
\end{IEEEbiography}

\begin{IEEEbiography}[{\includegraphics[width=1in,height=1.25in,clip,keepaspectratio]{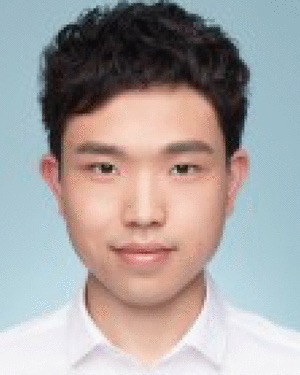}}]{Huan Li}
(Member, IEEE) received the PhD degree from Zhejiang University in 2018. He was an assistant professor at Aalborg University in Denmark from 2020 to 2023 and a senior engineer at Alibaba from 2018 to 2019. He is currently a ZJU100 professor at Zhejiang University and was awarded an EU Marie Curie Individual Fellowship. His work has appeared in top-tier journals and conferences. His research interests include data management for AI, spatio-temporal computing, and mobile computing.
\end{IEEEbiography}

\begin{IEEEbiography}[{\includegraphics[width=1in,height=1.25in,clip,keepaspectratio]{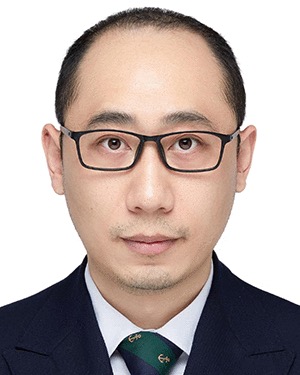}}]{Dalin Zhang}
(Senior Member, IEEE) received the BS degree from Jilin University in 2012, the MS degree from the University of Chinese Academy of Sciences in 2015, and the PhD degree from the University of New South Wales (UNSW Sydney) in 2020. He is currently a professor at Hangzhou Dianzi University, China. His research interests include brain--computer interfaces, human activity recognition, and the Internet of Things.
\end{IEEEbiography}

\begin{IEEEbiography}[{\includegraphics[width=1in,height=1.25in,clip,keepaspectratio]{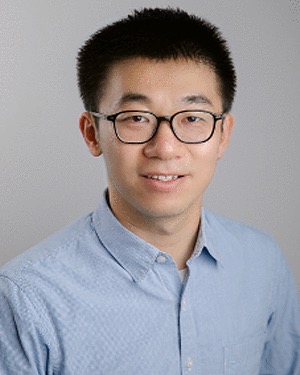}}]{Dong Gong}
is a Senior Lecturer and ARC DECRA Fellow at the School of Computer Science and Engineering, The University of New South Wales (UNSW Sydney), Australia. He also holds an adjunct position with the Australian Institute for Machine Learning (AIML) at The University of Adelaide. Before joining UNSW in 2022, he was a Research Fellow at AIML and a Principal Researcher at the Centre for Augmented Reasoning (CAR), The University of Adelaide.
\end{IEEEbiography}

\begin{IEEEbiography}[{\includegraphics[width=1in,height=1.25in,trim=0 126bp 0 0,clip,keepaspectratio]{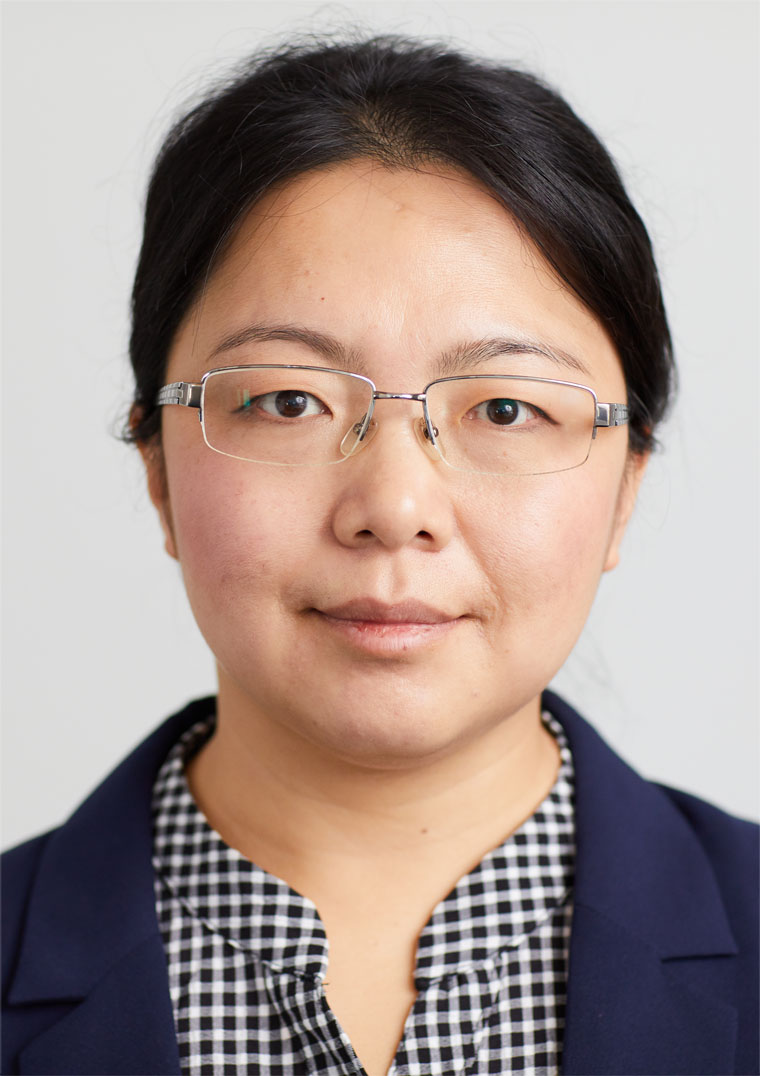}}]{Lina Yao}
(Senior Member, IEEE) received the master's and Ph.D. degrees in computer science from The University of Adelaide, Australia, in 2010 and 2014, respectively. She is a Professor at UNSW Sydney and an ARC Future Fellow, as well as a Visiting Scientist at CSIRO's Data61 and a Co-Theme Leader at the Responsible AI Research Centre. She has been a Clarivate Highly Cited Researcher since 2024. Her research interests include machine learning and data mining for the Internet of Things, recommendation, human activity recognition, brain--computer interfaces, and causal and responsible AI.
\end{IEEEbiography}

\begin{IEEEbiography}[{\includegraphics[width=1in,height=1.25in,clip,keepaspectratio]{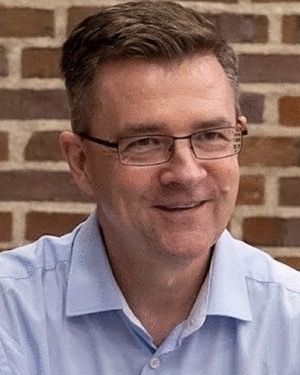}}]{Christian S. Jensen}
(Fellow, IEEE) received the PhD degree from Aalborg University in 1991 and the DrTechn degree from Aalborg University in 2000. He is currently a professor with the Department of Computer Science, Aalborg University. His research focuses primarily on temporal and spatio-temporal data management and analytics, including indexing and query processing, data mining, and machine learning.
\end{IEEEbiography}

\clearpage
\end{document}